\documentclass{article}

\usepackage[final]{nips_2017}

\usepackage[utf8]{inputenc} % allow utf-8 input
\usepackage[T1]{fontenc}    % use 8-bit T1 fonts
\usepackage{hyperref}       % hyperlinks
\usepackage{url}            % simple URL typesetting
\usepackage{booktabs}       % professional-quality tables
\usepackage{amsfonts}       % blackboard math symbols
\usepackage{nicefrac}       % compact symbols for 1/2, etc.
\usepackage{microtype}      % microtypography
\usepackage{enumerate}
\usepackage[usenames,dvipsnames]{color}
\usepackage{xcolor}
\usepackage[pdftex]{graphicx}
\usepackage{wrapfig}
\usepackage{paralist}
\usepackage{multirow}
\usepackage{amsmath,amsfonts,amssymb}
\usepackage{verbatim}
\usepackage{anyfontsize}

\newcommand{\ie}{\emph{i.e.},}

\newcommand{\eg}{\emph{e.g.},}
\newcommand\norm[1]{\left\lVert#1\right\rVert}

\newcommand\mytextsf[1]{{\fontsize{8.5}{10}\textsf{#1}}}

\graphicspath{{images/}}
\DeclareGraphicsExtensions{.png,.jpg,.pdf}

\title{A Deep Compositional Framework for Human-like Language
  Acquisition in Virtual Environment}

\author{
  Haonan Yu, Haichao Zhang, and Wei Xu\\
  Baidu Research - Institue of Deep Learning\\
  Sunnyvale, CA 94089\\
  \texttt{\{haonanyu,zhanghaichao,xuwei06\}@baidu.com}\\
}

\begin{document}

\abovedisplayskip=0pt
\abovedisplayshortskip=0pt
\belowdisplayskip=0pt
\belowdisplayshortskip=0pt

\maketitle

\begin{abstract}
  We tackle a task where an \emph{agent} learns to navigate in a 2D
  maze-like environment called \textsc{xworld}.
  In each session, the agent perceives a sequence of raw-pixel
  frames, a natural language command issued by a \emph{teacher}, and a
  set of rewards.
  The agent learns the teacher's language from scratch in a grounded
  and compositional manner, such that after training it is able to
  correctly execute \emph{zero-shot} commands: 1) the combination of
  words in the command never appeared before, and/or 2) the command
  contains \emph{new object concepts} that are learned from another
  task but never learned from navigation.
  Our deep framework for the agent is trained end to end: it learns
  simultaneously the visual representations of the environment, the
  syntax and semantics of the language, and the action module that
  outputs actions.
  The zero-shot learning capability of our framework results from
  its compositionality and modularity with parameter tying.
  We visualize the intermediate outputs of the framework,
  demonstrating that the agent truly understands how to solve the
  problem.
  We believe that our results provide some preliminary insights on
  how to train an agent with similar abilities in a 3D environment.
\end{abstract}

\section{Introduction}
\label{sec:introduction}
\vspace{-2ex}

\begin{wrapfigure}{r}{0.5\textwidth}
  \vspace{-15pt}
  \setlength{\tabcolsep}{0.5pt}
  \begin{center}
    \begin{tabular}{@{}cc@{}}
      \includegraphics[width=0.25\textwidth]{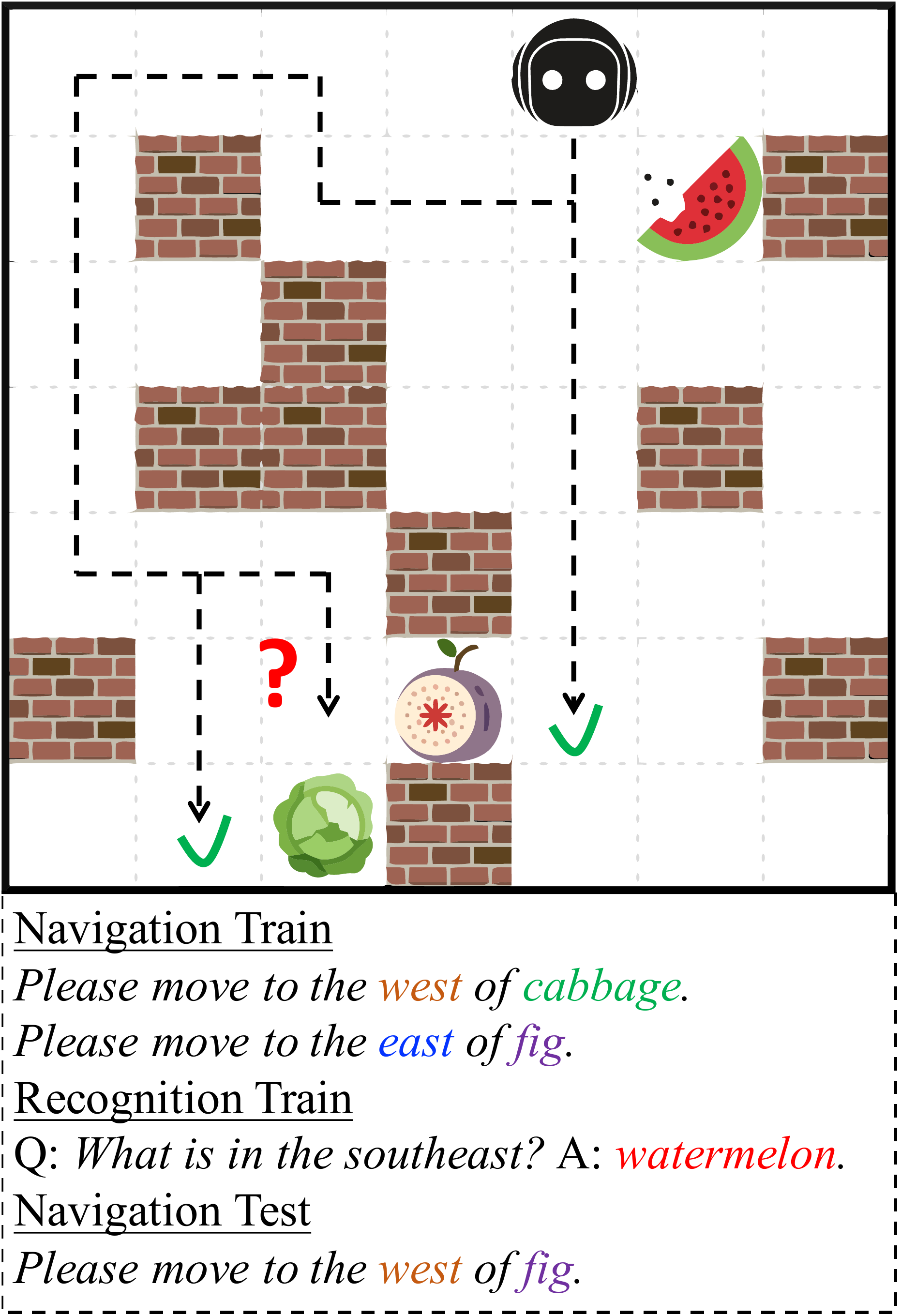} &
      \includegraphics[width=0.25\textwidth]{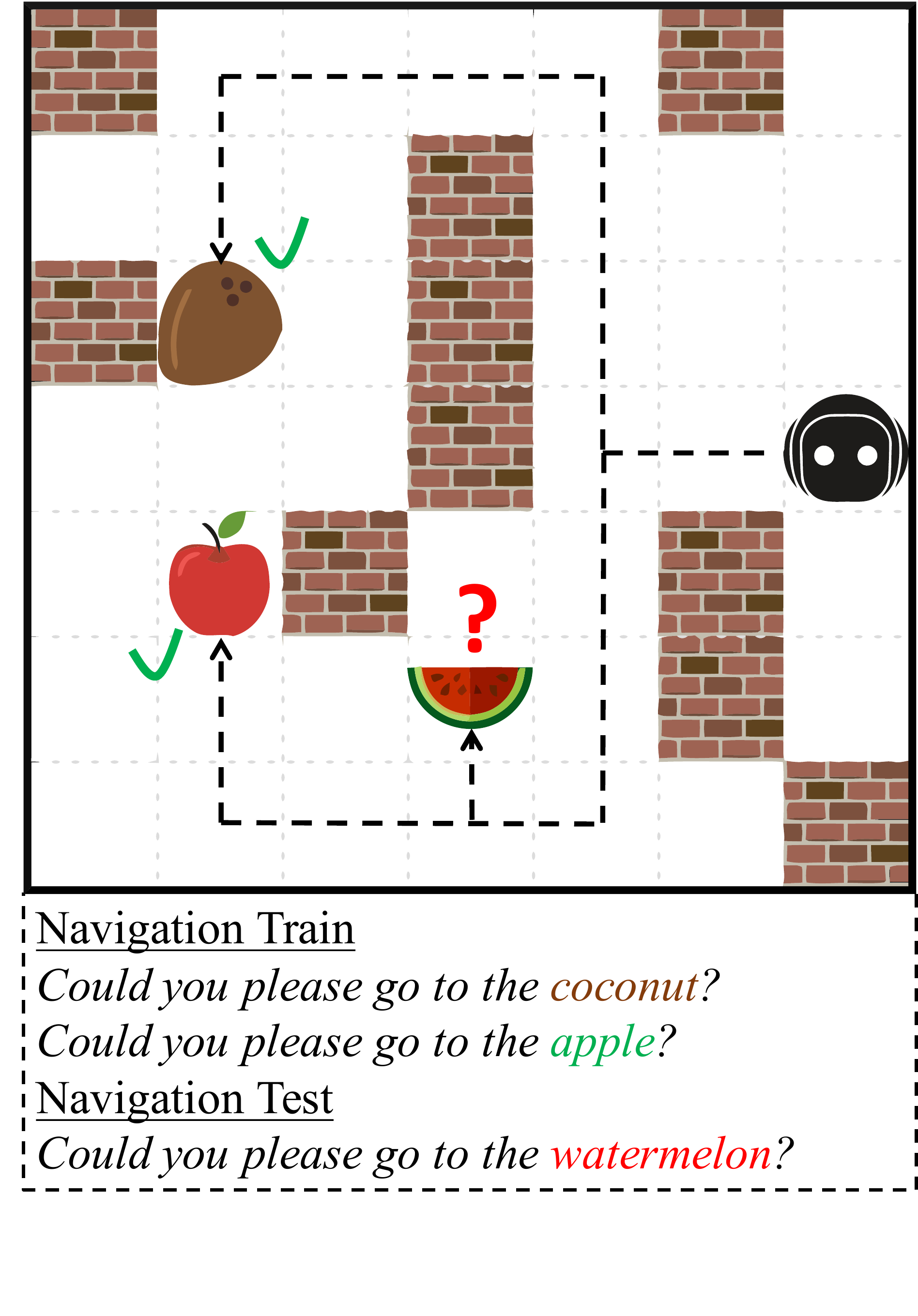}\\
      (a) & (b)\\
    \end{tabular}
    \caption{Illustration of our \textsc{xworld} environment and the
      zero-shot navigation tasks.
      (a) Test command contains an unseen word combination.
      (b)~Test command contains completely new object concepts that
      are learned from the recognition task in some previous sessions
      (a).
    }
    \label{fig:task}
  \end{center}
  \vspace{-20pt}
\end{wrapfigure}

The development of a sophisticated language system is a very crucial
part of achieving human-level intelligence for a machine.
Language semantics, when grounded in perception experience, can encode
knowledge about perceiving the world.
This knowledge is transferred from task to task, which empowers the
machine with generalization ability.
It is argued that a machine has to go through physical experience in
order to learn human-level semantics \citep{Kiela2016}, \ie\ a process
of \emph{human-like} language acquisition.
However, current machine learning techniques do not have a reasonably
fast learning rate to make this happen.
Thus we model this problem in a virtual environment, as the first step
towards training a physical intelligent machine.

Human generalize surprisingly well when learning new concepts and
skills through natural language instructions.
We are able to apply an existing skill to newly acquired concepts with
little difficulty.
For example, a person who has learned how to execute the command
``\textit{cut X with knife}'' when ``\textit{X}'' equals to
\textit{apple}, will do correctly when ``\textit{X}'' is something
else he knows (\eg\ \textit{pear}), even though he may have never been
asked to cut anything other than apple before.
In other words, human have the \emph{zero-shot} learning ability.

This paper describes a framework that demonstrates the zero-shot
learning ability of an agent in a specific task, namely, learning to
navigate in a 2D maze-like environment called \textsc{xworld}
(Figure~\ref{fig:task}).
We are interested in solving a similar task that is faced by a baby
who is learning to walk and navigate, at the stage of learning his
parents' language.
The parents might give some simple navigation command consisting of
only two or three words in the beginning, and gradually increase the
complexity of the command as time goes by.
Meanwhile, the parents might teach the baby the language in some other
task such as object recognition.
After the baby understands the language and masters the navigation
skill, he could immediately navigate to a new concept that is learned
from object recognition but never appeared in the navigation command
before.

We train our baby \emph{agent} across many learning sessions in
\textsc{xworld}.
In each session, the agent perceives the environment through a
sequence of raw-pixel images, a natural language command issued by a
\emph{teacher}, and a set of rewards.
The agent also occasionally receives the teacher's questions on object
recognition whenever certain conditions are triggered.
By exploring the environment, the agent learns simultaneously the
visual representations of the environment, the syntax and semantics of
the language, and how to navigate itself in the environment.
The whole framework employed by the agent is trained end to end from
scratch by gradient descent.
We test the agent under four different command conditions, three of
which require that the agent generalizes to interpret unseen commands
and words, and that the framework architecture is modular so that
other modules such as perception and action will still work properly
under such circumstance.
Our experiments show that the agent performs equally well ($\sim90\%$
average success rate) in all conditions.
Moreover, several baselines that simply learn a joint embedding for
image and language yield poor results.

In summary, our main contributions are two-fold:
\begin{compactenum}[$\circ$]
  \item A new deep reinforcement learning (DRL) task that integrates
    both vision and language.
    The language is \emph{not} pre-parsed \citep{Sukhbaatar2016} or
    -linked \citep{Mikolov2015,Sukhbaatar2016} to the environment.
    Instead, the agent has to learn everything from scratch and
    ground the language in vision.
    This task models a similar scenario faced by a learning child.
  \item The zero-shot learning ability by leveraging the
    compositionality of both the language and the network architecture.
    We believe that this ability is a crucial component of
    human-level intelligence.
\end{compactenum}

\vspace{-4ex}
\section{Related Work}
\vspace{-2ex}
Our work is inspired by the research of multiple disciplines.
Our \textsc{xworld} is similar to the MazeBase environment
\citep{Sukhbaatar2016} in that both are 2D rectangular grid world.
One big difference is that their quasi-natural language is already
parsed and linked to the environment.
They put more focus on reasoning and planning but not language
acquisition.
On the contrary, we emphasize on how to ground the language in vision
and generalize the ability of interpreting the language.
There are several challenging 3D environments for RL such as
ViZDoom \citep{Kempka2016} and DeepMind Lab \citep{Beattie2016}.
The visual perception problems posed by them are much more difficult
than ours.
However, these environments do not require language understanding.
Our agent needs to learn to interpret different goals from different
natural language commands.

Our setting of language learning shares some similar ideas of the AI
roadmap proposed by \citet{Mikolov2015}.
Like theirs, we also have a teacher in the environment that assigns
tasks and rewards to the agent.
The teacher also provides additional questions and answers to the
agent in an object recognition task.
Unlike their proposal of entirely using linguistic channels, our tasks
involve multiple modalities and are more similar to human experience.

The importance of compositionality and modularity of a learning
framework has been discussed at length in cognitive science by
\citet{Lake2016}.
The compositionality of our framework is inspired by the ideas in
Neural Programmer \citep{Neelakantan2016} and Neural Module
Networks (NMNs)\citep{Andreas2016a,Andreas2016b}.
Neural Programmer is trained with backpropagation by employing soft
operations on databases.
NMNs assemble several primitive modules according to questions in
Visual Question Answering (VQA).
It depends on an external parser to convert each sentence to one or
several candidate parse trees and thus cannot be trained end to end.
We adapt their primitive modules to our framework with differentiable
computation units to enable gradient calculation.
Their subsequent work (N2NMNs)~\citep{Hu2017} learns end to end to
optimize over the full space of parse trees, but relies on candidate
parse trees as an expert policy used by a behavior cloning procedure
to initialize the policy parameters.

Our recognition task is essentially image VQA
\citep{Antol2015,Gao2015,Ren2015,Lu2016,Andreas2016a,Andreas2016b,Teney2016,Yang2016}.
The navigation task can also be viewed as a VQA problem if the actions
are treated as answer labels.
Moreover, it is a zero-shot VQA problem (\ie\ test questions
containing unseen concepts) which has not been well addressed yet.

Our language acquisition problem is closely related to some recent
work on grounding language in images and videos
\citep{Yu2013,Rohrbach2016,Gao2016}.
The navigation task is also relevant to robotics navigation under
natural language command \citep{Chen2011,Tellex2011,Barrett2015}.
However, they either assume annotated navigation paths in the
training data or do not ground language in vision.
As \textsc{xworld} is a virtual environment, we currently do not
address mechanics problems encountered by a physical robot, but focus
on its mental model building.

\vspace{-2ex}
\section{\textsc{xworld} Environment}
\label{sec:environment}
\begin{figure}[t]
  \begin{center}
    \resizebox{\textwidth}{!}{
      \begin{tabular}{@{}c|c@{}}
        \includegraphics[width=0.55\textwidth]{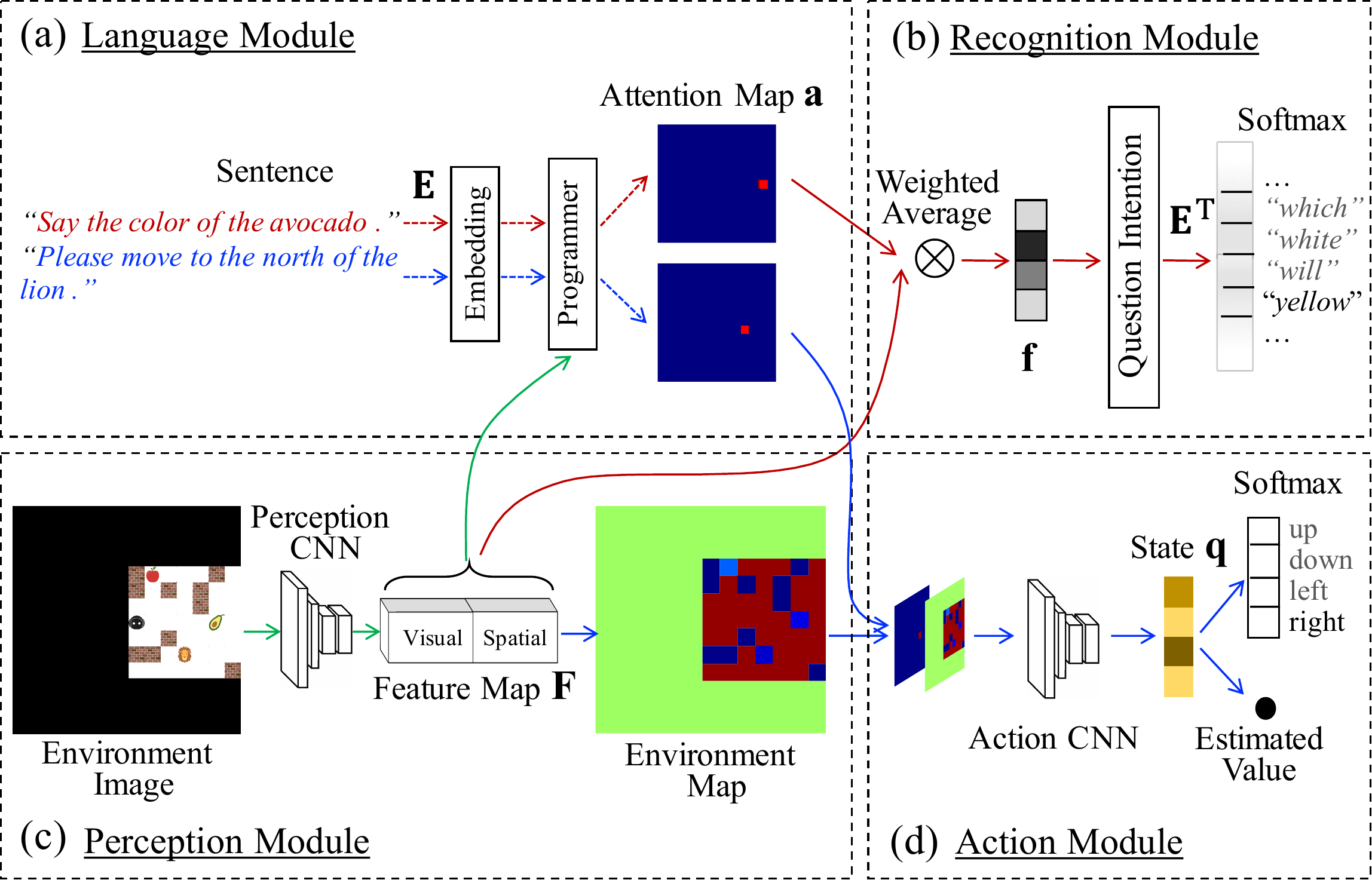}&
        \includegraphics[width=0.44\textwidth]{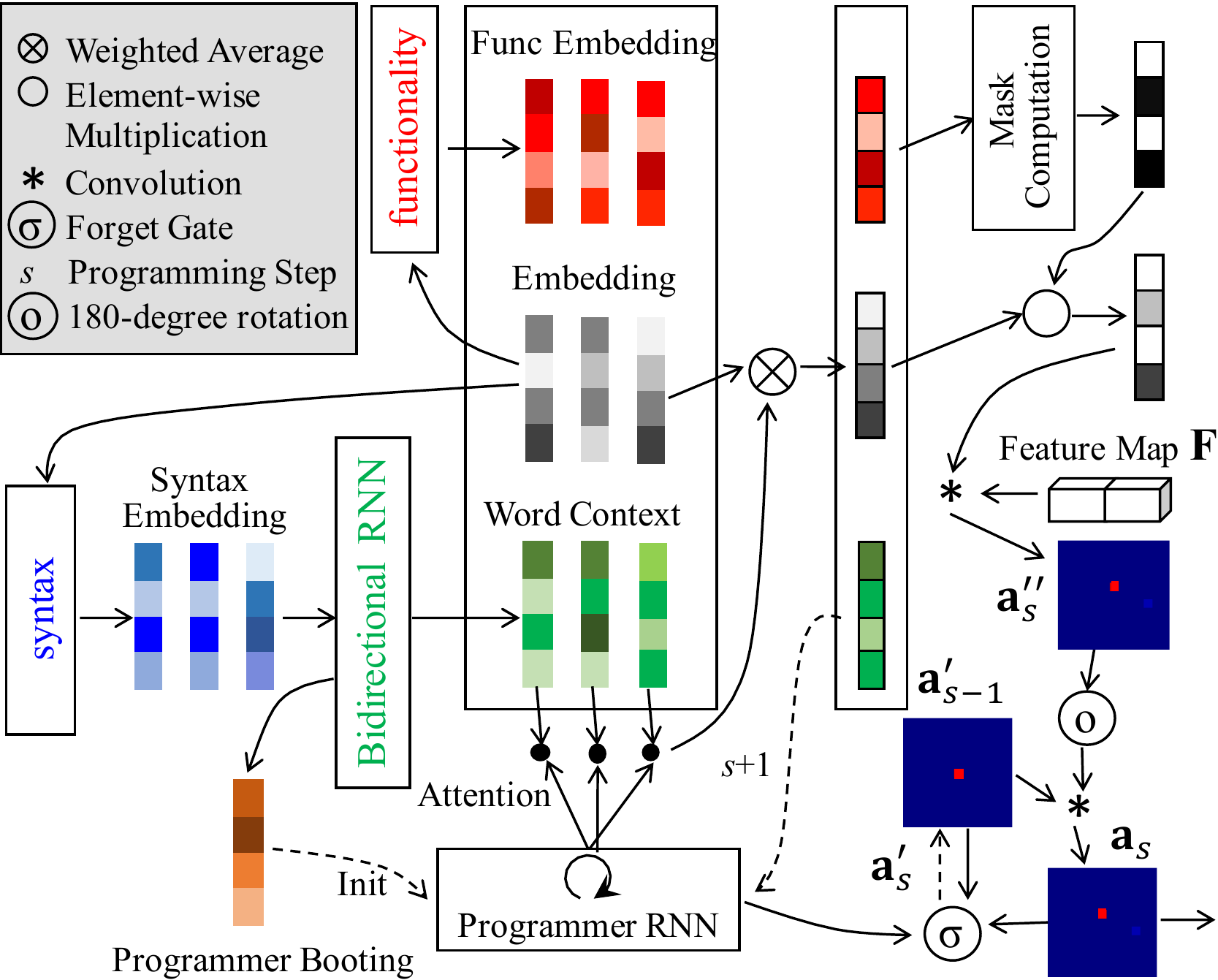}\\
      \end{tabular}
    }
    \caption{Left: An overview of our framework.
      The inputs are an environment image and a sentence (either a
      navigation {\color{blue}command} or a {\color{red}question}).
      The output is either a navigation {\color{blue}action} or an
      {\color{red}answer} to the question, respectively.
      The red and blue lines in (a) indicate different tasks going
      through exactly the same process.
      Right: The pipeline of the programmer of the language module.
      The input is a sequence of word embeddings.
      The output is the attention map at the final step.
    }
    \label{fig:overview}
  \end{center}
  \vspace{-3ex}
\end{figure}

\vspace{-2ex}
We first briefly describe the \textsc{xworld} environment.
More details are in Appendix~\ref{app:xworld}.
\textsc{xworld} is a 2D grid world (Figure~\ref{fig:task}).
An agent interacts with the environment over a number of time
steps~$T$, with four actions: \texttt{up}, \texttt{down},
\texttt{left}, and \texttt{right}.
It does so for many sessions.
At the beginning of each session, a teacher starts a timer and issues
a natural language command asking the agent to reach a location
referred to by objects in the environment.
There might be other objects as distractors.
Thus the agent needs to differentiate and navigate to the right
location.
It perceives the entire environment through RGB pixels with an
egocentric view (Figure~\ref{fig:overview}c).
If the agent correctly executes the command before running out of
time, it gets a positive reward.
Whenever it hits a wall or steps on an object that is not the target,
it gets a negative reward.
The agent also receives a small negative reward at every step as a
punishment for taking too much time.
After each session, the environment is reset randomly.

Some example commands are (the parentheses contain environment
configurations that are \emph{withheld} from the agent, same below):
\begin{compactenum}[$\circ$]
\item \textit{Please navigate to the apple.}
  (There is an apple, a banana, an orange, and a grape.)
\item \textit{Can you move to the grid between the apple and the banana?}
  (There is an apple and a banana.
  The apple and the banana are separated by one empty grid.)
\item \textit{Could you please go to the red apple?}
  (There is a green apple, a red apple, and a red cherry.)
\end{compactenum}

The difficulty of this navigation task is that, at the very beginning
the agent knows nothing about the language: every word appears equally
meaningless.
After trials and errors, it has to figure out the language syntax and
semantics in order to correctly execute the command.

While the agent is exploring the environment, the teacher also asks
object-related questions whenever certain conditions are triggered
(all conditions are listed in Appendix~\ref{app:xworld}).
The answers are always single words and provided by the teacher for
supervision.
Some example QA pairs are:
\begin{compactenum}[$\circ$]
\item Q:\textit{What is the object in the north?}
  A:\textit{Banana.}
  (The agent is by the south of a banana, by the north of an apple,
  and by the west of a cucumber.)
\item Q:\textit{Where is the banana?}
  A:\textit{North.}
  (The agent is by the south of a banana and the east of an apple.)
\item Q:\textit{What is the color of the object in the west of the
  apple?}
  A:\textit{Yellow.}
  (An apple has a banana on its west and a cucumber on its east.)
\end{compactenum}
We expect the agent to transfer the knowledge exclusively learned
from this recognition task to the navigation task to execute zero-shot
commands.
Both the commands and questions are generated from templates
(Appendix~\ref{app:xworld}) and are made like human-elicited
sentences.

\vspace{-2ex}
\section{Compositional Framework for Zero-shot Navigation}
\label{sec:framework}
\vspace{-1ex}
The zero-shot navigation problem can be formulated as training a
policy with a grounding model:
\[\pi(y|\mathcal{E},\mathbf{a}), \text{with}~\mathbf{a}=Grounding(\mathcal{E},C)\]
that works correctly for a new test command $C'$ which never appears in
the training set.
In the above formulation, $y$ is the agent's navigation action,
$\mathcal{E}$ is the perceived environment, $\mathbf{a}$ is the
navigation goal attention computed by grounding the teacher's command
$C$ in the environment $\mathcal{E}$.
Note that the policy $\pi(y|\mathcal{E},\mathbf{a})$ depends on
$\mathbf{a}$ but not directly on $C$, thus our key problem is
essentially how to ground a new command to get the correct attention
map.

The proposed framework for solving this problem contains four major
modules: perception (Section~\ref{sec:perception}), recognition
(Section~\ref{sec:recognition}), language
(Section~\ref{sec:language}), and action (Section~\ref{sec:action}).
The design of the framework is driven by the need of navigating under
new command $C'$.
Thus we will focus on how language grounding
(Figure~\ref{fig:overview}a) and recognition
(Figure~\ref{fig:overview}b) are related, and how to transfer the
knowledge learned in the latter to the former.
For an effective knowledge transfer, we believe that the framework has
to possess three crucial properties:
\begin{compactenum}
\item[\textbf{PI}] Language grounding and recognition have to be
  reduced to (approximately) the \emph{same} problem.
  This ensures that a new word exclusively learned from
  recognition has already been grounded for attention.
\item[\textbf{PII}] The language module must be compositional.
  It must process a sentence while preserving the (major) sentence
  structure.
  One example would be a parser that outputs a parse tree.
\item[\textbf{PIII}] Inductive bias \citep{Lake2016} must be learned
  from training sentences.
  The language module must know how to parse a sentence that has a
  previously seen structure even if a word position of the sentence is
  filled with a completely new word.
\end{compactenum}

There is no existing framework for image captioning or VQA exhibits
all such properties.
While in theory, binding parameters or embeddings between the language
module and the recognition module might satisfy the first property, in
practice a naive binding without further architecture support hardly
works.
It is also challenging for a simple (gated) Recurrent Neural Network
(RNN) to satisfy the last two properties when a sentence has a rich
structure.
Our framework detailed below is designed exactly for these three
properties.

\vspace{-2ex}
\subsection{Generating Feature Map}
\label{sec:perception}
\vspace{-2ex}
We compute a feature map $\mathbf{F}$ as an abstraction of the
environment and the agent's ego-centric geometry, on which we perform
the recognition and grounding of the \emph{entire} dictionary.
Because of this general computation purpose, the map should not only
encode visual information, but also provide spatial features.
The feature map is formed by stacking two feature maps $\mathbf{F}_v$
and $\mathbf{F}_s$ along the channel dimension.
The perception module (Figure~\ref{fig:overview}c) transforms an
environment image into the visual feature map $\mathbf{F}_v$ through a
Convolutional Neural Network (CNN), whose output has a spatial
dimension equal to the number of grids $N$ in the environment, where
each pixel corresponds to a grid.
The spatial feature map $\mathbf{F}_s$ is a set of randomly
initialized parameters.
Intuitively, $\mathbf{F}_v$ is responsible for the agent's visual
sense while $\mathbf{F}_s$ is responsible for its direction sense.
The agent needs to learn to associate visually-related words with
$\mathbf{F}_v$ while associate spatially-related words with
$\mathbf{F}_s$.

\vspace{-2ex}
\subsection{Recognizing and Grounding a Single Word}
\label{sec:recognition}
\vspace{-1ex}
To achieve zero-shot navigation, we start with reducing the
single-word recognition and grounding to the same problem, after which
we will extend the reduction to the grounding of a sentence of
multiple words (Section~\ref{sec:language}).
The single-word grounding problem is defined as
\begin{equation}
  \label{eq:grounding}
  \mathbf{F}, m \mapsto \mathbf{a},
\end{equation}
where $\mathbf{F}\in\mathbb{R}^{D\times N}$ is a feature map, $m$ is
the word index, $\mathbf{a}\in\mathbb{R}^N$ is a spatial attention
map, and $N$ is the number of pixels.
Imagine that the feature map is a collection of slots of features,
given a word $m$, we need to find the corresponding slots (highlighted
by attention) that match the word semantics.
In a reverse way, the recognition problem is defined as
\begin{equation*}
  \mathbf{F}, \mathbf{a} \mapsto m.
\end{equation*}
The recognition module (Figure~\ref{fig:overview}b) outputs the most
possible word for a feature map weighted by attention.
This reflects the question answering process after the agent attending
to a region of the environment image according to a question.
In the extreme case, when $\mathbf{a}$ is a one-hot vector, the module
extracts and classifies one slot of feature.
Suppose that the extracted feature $\textbf{f}=\textbf{F}\textbf{a}$
is directly fed to a Softmax layer
$P(m|\mathbf{f})=Softmax_M(\mathbf{s}_m^{\intercal}\mathbf{f})=Softmax_M((\mathbf{s}_m^{\intercal}\mathbf{F})\mathbf{a})$,
where $Softmax_M$ denotes the Softmax function over $M$ classes and
$\mathbf{s}_m$ is the parameter vector of the $m$th class.
Notice that the multiplication $\mathbf{s}_m^{\intercal}\mathbf{F}$
essentially treats $\mathbf{s}_m$ as a $1\times1$ filter and convolves
it with every feature on $\mathbf{F}$.
This computation operates as if it is solving the grounding problem in
Eq.~\ref{eq:grounding}, and it produces an attention map
$\mathbf{s}_m^{\intercal}\mathbf{F}$ which is optimized towards the
following property
\begin{equation*}
  \mathbf{s}_m^{\intercal}\mathbf{F}\propto\left\{
    \begin{array}{ll}
      \mathbf{a} & m=m^*\\
      -\mathbf{a} & m\neq m^*
    \end{array},
    \right.
\end{equation*}
where $m^*$ is the groundtruth class for $\mathbf{f}$.
In other words, this property implies that $\mathbf{s}_m$ is grounded in
$\mathbf{F}$ where the feature of class $m$ locates.
This is exactly where we need to ground word $m$ with embedding
$\mathbf{e}_m$.
Thus we set the Softmax matrix as the transpose of the word embedding
table (\ie\ $\mathbf{s}_m=\mathbf{e}_m$).
So far, we have satisfied property \textbf{PI} for the single-word
grounding case, namely, if a new word $M+1$ is trained to be
recognized accurately, then it can also be grounded correctly without
being trained to be grounded before.

There is still one flaw in the above definition of recognition.
An attention map only tells the agent \emph{where}, but not
\emph{what}, to focus on.
Imagine a scenario in which the agent stands to the east of a red apple.
Given an attention map that highlights the agent's west, should the
agent answer \textit{west}, \textit{apple}, or \textit{red}?
It turns out that the answer should also depend on the question
itself, \ie\ the question intention has to be understood.
Therefore, we redefine the recognition problem as:
\[\mathbf{F}, \mathbf{a}, \Phi(Q) \mapsto m,\]
where $Q$ is the question and $\Phi$ computes its intention.
In our framework, $\Phi$ is modeled as a gated RNN \citep{Cho2014}
which outputs an embedding mask $\in[0,1]^D$ through Sigmoid.
Then the Softmax distribution becomes
$P(m|\mathbf{f})=Softmax_M((\mathbf{e}_m\circ\Phi(Q))^{\intercal}\mathbf{F}\mathbf{a})=Softmax_M(\mathbf{e}_m^{\intercal}(\Phi(Q)\circ\mathbf{F}\mathbf{a}))$
(Figure~\ref{fig:overview}b), where~$\circ$ denotes element-wise
multiplication.
To make property \textbf{PI} still possible, we modify the grounding
problem so that a word embedding is masked before it is grounded:
\begin{equation}
  \label{eq:grounding-redefined}
  \mathbf{F}, m, \Psi(m) \mapsto \mathbf{a},
\end{equation}
where $\Psi$ is modeled as a projection\footnote{We use ``projection''
  to denote the process of going through one or more fully-connected
  (FC) layers.} that outputs an embedding mask through Sigmoid.
Intuitively, it enhances the grounding with the ability of selecting
features that are relevant to the word being grounded, and this
selection is only decided by the word itself.
Both functions $\Phi$ and $\Psi$ serve as the agent's decision makers
on which features to focus on given particular language tokens.
They operate on the same intermediate word representation referred to
as \emph{functionality embedding} that is a projection of the word
embedding (Figure~\ref{fig:projections} Appendix~\ref{app:figures}).
Functionality embedding is exclusively used for computing embedding masks.
Details of mask computation are in Figure~\ref{fig:masks}
Appendix~\ref{app:figures}.

\vspace{-2ex}
\subsection{Grounding a Sentence}
\label{sec:language}
\vspace{-1ex}
Now we extend single-word grounding to sentence grounding while trying
to preserve property \textbf{PI}.
The high-level idea is that we can ground a sentence by grounding
individual words of that sentence through several steps, one
or more words per step at a time.
The questions we will answer in this section are how to: 1) decide
which words to ground at each step, 2) combine the grounding results
of these steps to reflect the grounding of the sentence, and 3) make
the whole process differentiable.

The core of our language module (Figure~\ref{fig:overview}a) is a
\emph{differentiable programmer} (Figure~\ref{fig:overview} Right).
We treat each sentence (either navigation command or recognition
question) as a \emph{program command}.
The programmer converts the command implicitly to a set of operations
(see Appendix~\ref{app:operations} for an analogy between the
operations of NMNs and ours).
The programming result is an attention map.
For navigation, the attended regions indicate the target locations.
For recognition, they indicate which part of the environment to be
recognized.
The programmer programs a command in several steps.
At each step, the programming consists of three substeps: attending,
grounding, and combining.
The programmer first softly attends to a portion of a sentence.
Once one or more words are attended, they are extracted, modulated by
the embedding mask, and grounded in image.
The resulting attention map is selectively combined with the maps in
the previous steps (detailed below).
In this way, the programmer uses the dynamic word attention, the
grounding process modulated by masks, and the selective map
combination to implicitly output a network layout and model the
structural process of grounding a sentence in image.
This approximately satisfies property \textbf{PII}.

The word attention is determined by the global sentential context, the
local context at each word, and the history of a \emph{programmer
  RNN}.
Both global and local contexts are computed based on an intermediate
word representation referred to as \emph{syntax embedding} that is a
projection of the word embedding (Figure~\ref{fig:projections}
Appendix~\ref{app:figures}).
Intuitively, the syntax embedding can be trained to encode a group of
words similarly if they have the same syntactic meaning.
Thus when a sentence contains a new word that was only trained in
recognition, the programmer RNN has the inductive bias of interpreting
the sentence as if the new word is an existing word with the same
syntactic meaning (property \textbf{PIII}).
The details of word attention are in Appendix~\ref{app:word-attention}
and omitted here due to page limit.

With the word attention at each step $s$, we compute the averaged word
embedding, mask it given the corresponding averaged functionality
embedding, and convolve it with the feature map as in the single-word
case (Eq.~\ref{eq:grounding-redefined}).
The convolution result is input to a Softmax layer to get a
sum-to-one attention map $\mathbf{a}''_s$.
Assume that we cached an attention map $\mathbf{a}'_{s-1}$ in the
previous step.
The programmer approximates the 2D translation of spatial attention by
$\mathbf{a}_s=o(\mathbf{a}''_s)*\mathbf{a}'_{s-1}$, where $o$ denotes
the 180-degree rotation.
Then the programmer caches a new attention map through a forget gate:
$\mathbf{a}'_s=(1-\sigma)\mathbf{a}'_{s-1} + \sigma\mathbf{a}_s$,
where the gate $\sigma\in[0,1]$ is computed from the current
state of the programmer RNN (Figure~\ref{fig:overview} Right).
We set $\mathbf{a}_0'=\mathbf{i}$ where $\mathbf{i}$ is a map whose
center pixel is one and the rest are all zeros.
Finally, the attention map $\mathbf{a}$ at the last step is used as
the output of the programmer.

\vspace{-2ex}
\subsection{Navigation}
\label{sec:action}
\vspace{-1ex}
A navigation attention map $\mathbf{a}$ only defines the target
locations.
To navigate, the agent needs to know the surrounding environment
including obstacles and distractors.
Our perception module (Figure~\ref{fig:overview}c) convolves
$\mathbf{F}_v$ with a $1\times 1$ filter to get the environment map.
We stack the two maps and input them to an action CNN whose output
is projected to a state vector $\mathbf{q}$ that summarizes the
environment.
The state vector is further projected to a distribution
$\pi(\mathbf{q},y)$ over the actions.
At each time step, the agent takes action $y$ with a probability of
$\alpha\cdot 0.25 + (1-\alpha)\cdot\pi(\mathbf{q},y)$,
where $\alpha$ is the rate of random exploration.
The state is also projected to a scalar $V(\mathbf{q})$ to approximate
value function.

\vspace{-2ex}
\section{Training}
\vspace{-2ex}
Our training objective contains two sub-objectives
$\mathcal{L}(\theta)=\mathcal{L}_{RL}(\theta)+\mathcal{L}_{SL}(\theta)$,
one for navigation and the other for recognition, where $\theta$ are
the joint parameters of the framework.
Most parameters are shared between the two tasks\footnote{In all the
  figures of our framework, components with the same name share the
  same set of parameters.}.
We model the recognition loss $\mathcal{L}_{SL}$ as the multi-class
cross entropy which has the gradients
\begin{equation*}
  \nabla_{\theta}\mathcal{L}_{SL}(\theta)
  =\mathbb{E}_{Q}\left[-\nabla_{\theta}\log P_{\theta}(m|\mathbf{f}_{\theta})\right],
\end{equation*}
where $\mathbb{E}_{Q}$ is the expectation over all the questions asked
by the teacher in all training sessions, $m$ is the correct answer to
each question, and $\mathbf{f}_{\theta}$ is the corresponding
feature.
We compute the navigation loss $\mathcal{L}_{RL}(\theta)$ as the
negative expected reward $-\mathbb{E}_{\pi_{\theta}}[r]$ the agent
receives by following its policy $\pi_{\theta}$.
With the Actor-Critic (AC) algorithm \citep{Sutton1998}, we have the
approximate gradients
\begin{equation*}
  \nabla_{\theta}\mathcal{L}_{RL}(\theta)=-\mathbb{E}_{\pi_{\theta}}
  \left[\left(\nabla_{\theta}\log\pi_{\theta}(\mathbf{q}_{\theta},y)
    +\nabla_{\theta}V_{\theta}(\mathbf{q}_{\theta})\right)(r+\gamma
    V_{\theta^-}(\mathbf{q}_{\theta^-})-V_{\theta}(\mathbf{q}_{\theta}))\right]
\end{equation*}
where $\theta^-$ are the target parameters that are periodically
(every $J$ minibatches) copied from $\theta$, $r$ is the immediate
reward, $\gamma$ is the discount factor, $\mathbf{q}_{\theta^-}$ is
the next state after taking action $y$ at state $\mathbf{q}_{\theta}$,
and $\pi_{\theta}$ and $V_{\theta}$ are the policy and value output by
the action module.
Since the expectations $\mathbb{E}_Q$ and $\mathbb{E}_{\pi_{\theta}}$
are different, we optimize the two sub-objectives separately over the
same number of minibatches.
For effective training, we employ Curriculum Learning
\citep{Bengio2009} and Experience Replay \citep{Mnih2015} with
Prioritized Sampling \citep{Schaul2015} (Appendix~\ref{app:training}).

\vspace{-2ex}
\section{Experiments}
\vspace{-2ex}
We use Adagrad \citep{Duchi2011} with a learning rate of $10^{-5}$ for
Stochastic Gradient Descent (SGD).
In all experiments, we set the batch size to 16 and train 200k batches.
The target parameters $\theta^-$ are updated every $J=2\text{k}$
batches.
The reward discount factor~$\gamma$ is set to $0.99$.
All the parameters have a default weight decay equal to
$10^{-4}\times$ batch size.
For each layer, by default its parameters have zero mean and a
standard deviation of $1 \mathbin{/} \sqrt{K}$, where $K$ is the
number of parameters of that layer.
The agent has 500k exploration steps in total, and the exploration
rate $\alpha$ decreases linearly from $1$ to $0$.
We fix the number of programming steps as 3, and train each model with
4 random initializations.
The whole framework is implemented with PaddlePaddle
(\url{https://github.com/PaddlePaddle/Paddle/}) and trained end to
end.
More implementation details are in Appendix~\ref{app:implementation}.

\begin{figure}[h!]
  \begin{center}
    \includegraphics[width=\textwidth]{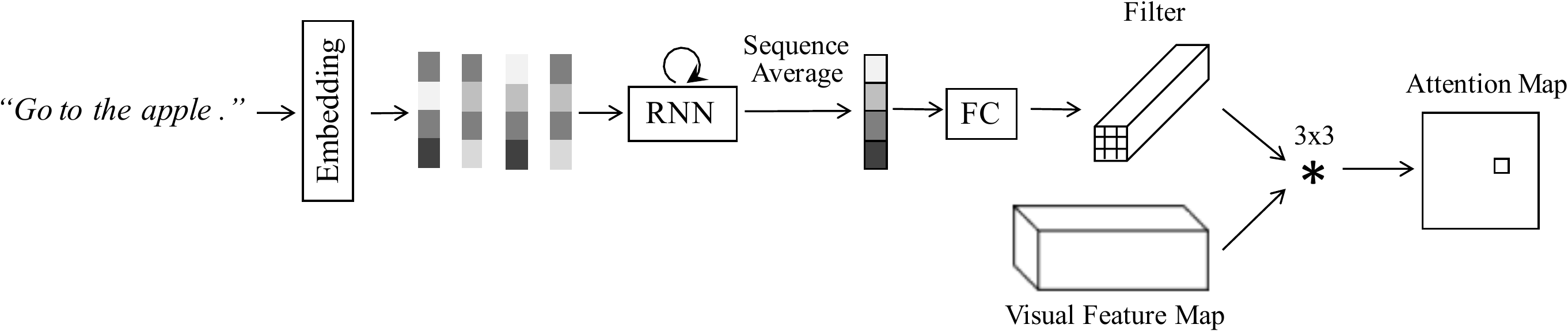}
    \caption{The pipeline of the language module of the
      \mytextsf{SimpleAttention} baseline.
      An RNN processes the input sentence and outputs a $3\times 3$
      filter which is convolved with the visual feature map generated
      by the perception module.
      Other modules are the same with our framework.
    }
    \label{fig:simple-attention}
  \end{center}
\end{figure}

\begin{figure}[h!]
  \begin{center}
    \includegraphics[width=0.9\textwidth]{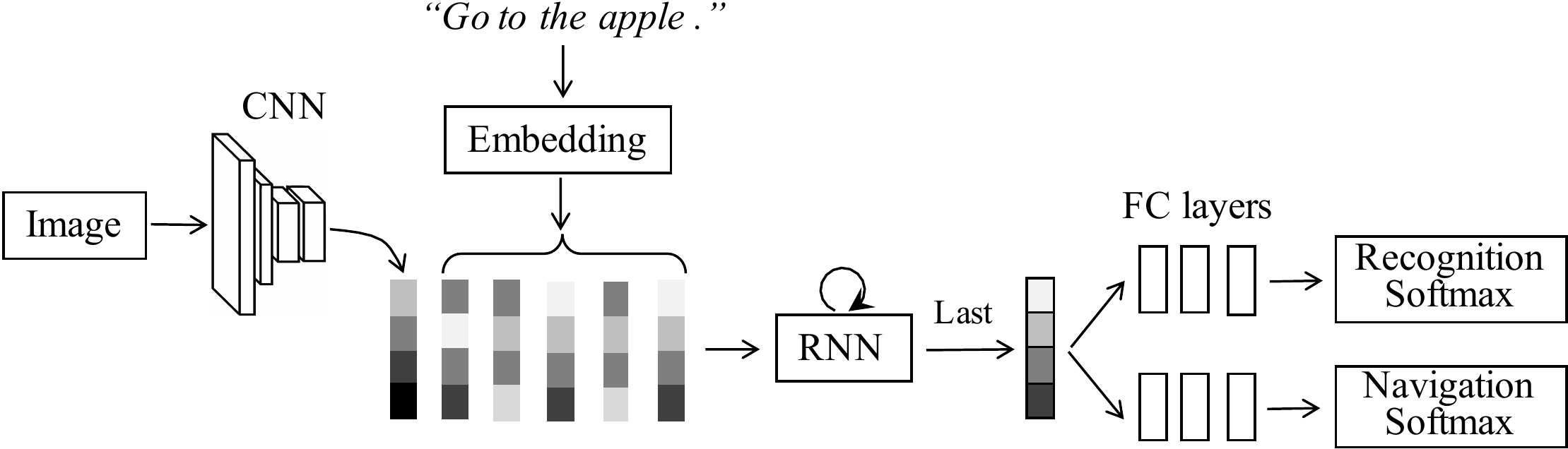}
    \caption{Our adapted version of the VIS-LSTM model \citep{Ren2015}.
      The framework treats the projected image embedding as the first
      word of the sentence.
      Navigation and recognition only differ in the last several FC
      layers.
    }
    \label{fig:vis-lstm}
  \end{center}
\end{figure}

\begin{figure}[h!]
  \begin{center}
    \includegraphics[width=\textwidth]{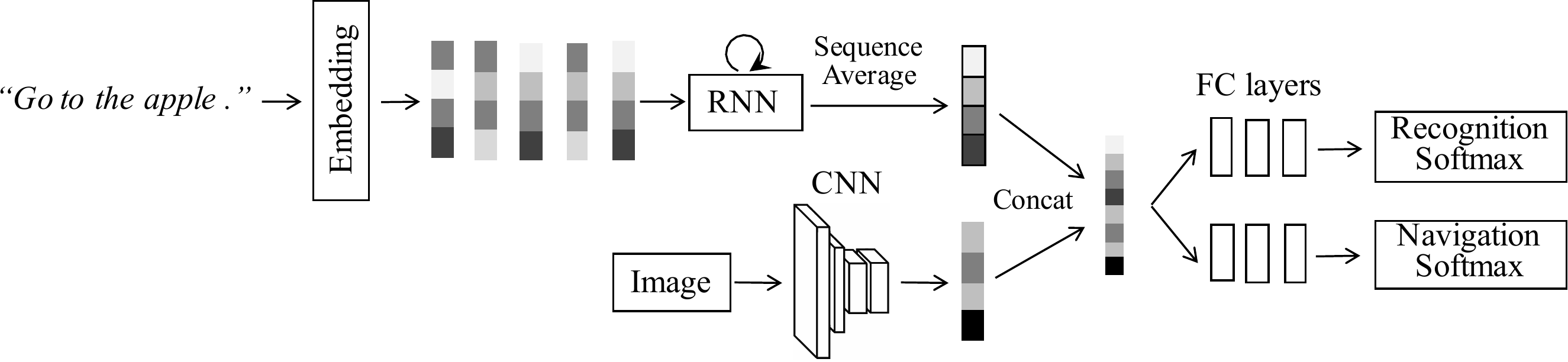}
    \caption{A multimodal framework adapted from the one in
      \citet{Mao2015}.
      The image and the sentence are summarized by a CNN and an RNN
      respectively.
      Their embeddings are concatenated and projected to the same
      space.
      Navigation and recognition only differ in the last several FC
      layers.
    }
    \label{fig:multimodal}
  \end{center}
\end{figure}

\textbf{Baselines} We first compare with five baselines in a normal
setting.
\begin{compactenum}[$\circ$]
\item{\mytextsf{SimpleAttention}} We modify our framework by replacing
  the programmer with a simple attention model.
  Given a sentence, we use an RNN to output an embedding which is
  convolved as a $3\times 3$ filter with the visual feature map to get
  an attention map.
  The rest of our framework is unchanged
  (Figure~\ref{fig:simple-attention} Appendix~\ref{app:figures}).
  This baseline is an ablation to show the necessity of the programmer.
\item{\mytextsf{NoTransShare}} We do not tie the word embedding table
  to the transposed Softmax matrix.
  In this case, language grounding and recognition are not the same
  problem.
  This baseline is an ablation to show the impact of the transposed
  sharing on the training convergence.
\item{\mytextsf{VIS-LSTM}} Following \citet{Ren2015}, we use CNN to
  get an image embedding which is then projected to the word embedding
  space and used as the first word of the sentence.
  The sentence goes through an RNN whose last state is used for
  navigation and recognition (Figure~\ref{fig:vis-lstm}
  Appendix~\ref{app:figures}).
\item{\mytextsf{Multimodal}} We implement a multimodal framework
  \citep{Mao2015}.
  This framework uses CNN to get an image embedding and RNN to
  get a sentence embedding.
  Then the two embeddings are projected to a common feature space for
  navigation and recognition (Figure~\ref{fig:multimodal}
  Appendix~\ref{app:figures}).
\item{\mytextsf{SAN}} We replace our language module with the
  attention process of Stacked Attention Network
  (SAN)~\citep{Yang2016}, with the difference of training a CNN from
  scratch, instead of using a pretrained one, to accommodate to
  \textsc{xworld}.
  The rest of our framework is unchanged.
\end{compactenum}

\begin{wrapfigure}{r}{0.5\textwidth}
  \vspace{-15pt}
  \begin{center}
    \begin{tabular}{cc}
      \includegraphics[width=0.45\textwidth]{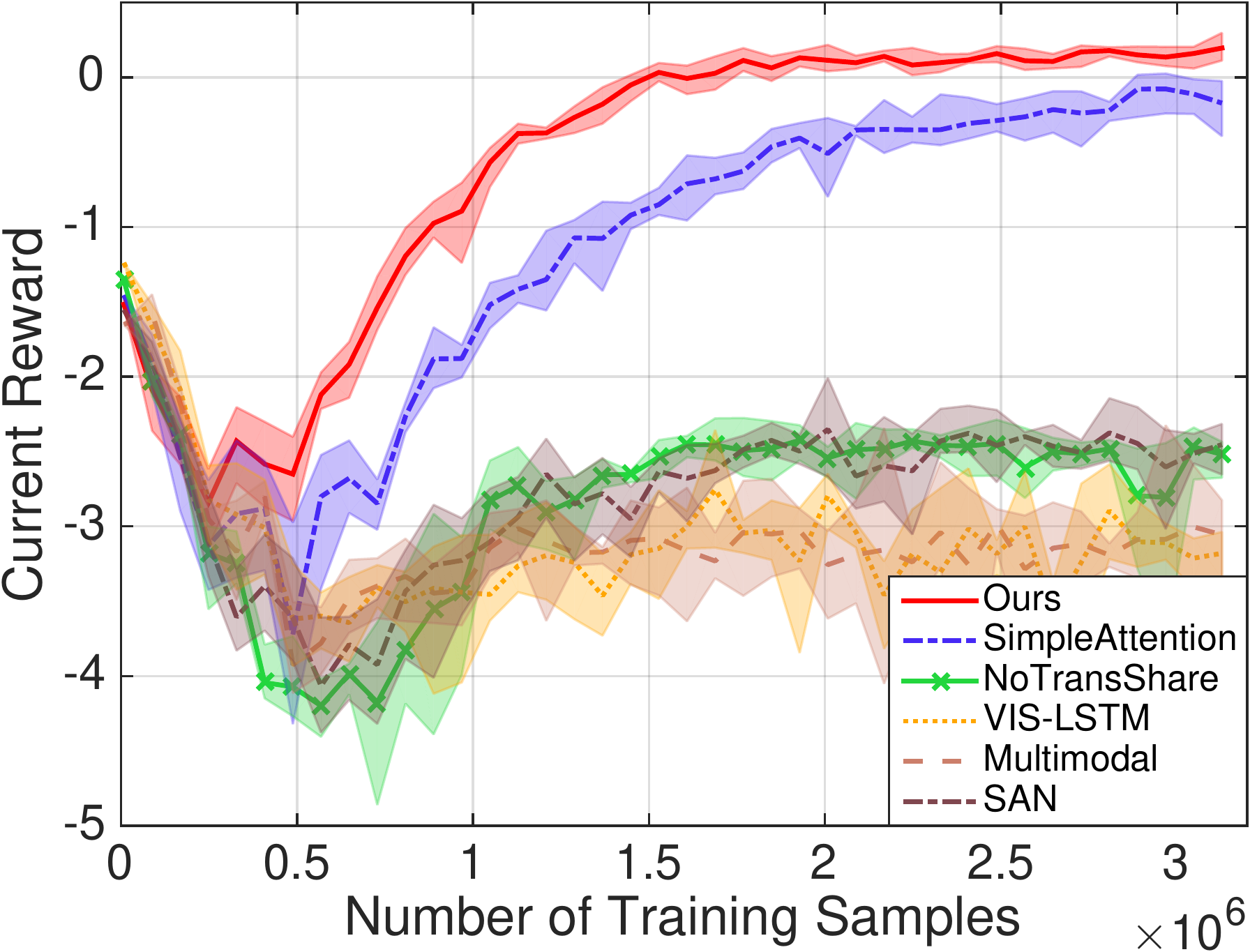}
    \end{tabular}
    \caption{Training reward curves of the five baselines and our
      framework.
      The shown reward is the accumulated discounted reward per
      session, averaged every 8k training examples.
      The shaded area of each curve denotes the variance among 4
      random initializations.
    }
    \label{fig:rewards}
  \end{center}
  \vspace{-10pt}
\end{wrapfigure}

For each method, we test 10k sessions in total over four navigation
subtasks \texttt{nav\_obj}, \texttt{nav\_col\_obj},
\texttt{nav\_nr\_obj}, and \texttt{nav\_bw\_obj}
(Appendix~\ref{app:xworld}).
We compute the success rates where success means that the agent
reaches the target location in time in a test session.
The training reward curves and the success rates are shown in
Figure~\ref{fig:rewards} and Table~\ref{tab:rates}a, respectively.
\mytextsf{VIS-LSTM} and \mytextsf{Multimodal} have poor results
because they do not ground language in vision.
\mytextsf{SAN} has the same overall architecture with our framework
but with a different language module, which requires it to generate
accurate navigation attention map because the navigation action solely
depends on attention maps rather than visual image features as in
\mytextsf{VIS-LSTM} and \mytextsf{Multimodal}.
This assumption is a quite strong one and our experiment result shows
that even though \mytextsf{SAN} gets $\sim$100\% recognition accuracy,
its action module hardly works.
Our explanation is that \mytextsf{SAN} reasons spatial relationships
based on the encoding of multiple neighboring objects in a single
feature at each convolution location (\ie\ a broad receptive field),
which can easily lead to attention diffusion.
Its recognition module might be able to still correctly classify given
diffused attention, but the action module requires far more accurate
attention as the navigation signal for the agent.
Surprisingly, \mytextsf{NoTransShare} converges much slower than our
framework does.
One possible reason is that the correct behavior of the language
module is hard to be found by SGD if no constraint on word embedding
is imposed.
Although \mytextsf{SimpleAttention} is able to perform well, without
word-level attention, its ability of sentence understanding is
limited.

\textbf{Zero-shot Navigation} Our primary question is whether the
agent has the zero-shot navigation ability of executing previously
unseen commands.
We setup four command conditions for training the agent:
\begin{compactenum}[$\circ$]
\item{\mytextsf{Standard}} The training command set has the same
  distribution with the test command set.
\item{\mytextsf{NC}} Some word combinations are excluded from
  the training command set, even though all the words are in it.
  We specifically consider three types of word combinations:
  (object, location), (object, color), and (object, object).
  We enumerate all combinations for each type and randomly hold out
  10\% from the teacher in navigation.
\item{\mytextsf{NWNav}\&\mytextsf{NWNavRec}} Some object words are
  excluded from navigation training, and are trained only in
  recognition and tested in navigation as new concepts.
  \mytextsf{NWNavRec} guarantees that the new words will not appear
  in questions but only in answers while \mytextsf{NWNav} does not.
  We randomly hold out 10\% of the object words.
\end{compactenum}

\begin{wrapfigure}{r}{0.43\textwidth}
  \setlength{\tabcolsep}{1.5pt}
  \centering
  \resizebox{0.43\textwidth}{!}{
    \begin{tabular}{@{}c@{}}
      \begin{tabular}{@{}c|c|c|c|c|c@{}}
        Ours & \mytextsf{SimpleAttention} & \mytextsf{NoTransShare} &
        \mytextsf{VIS-LSTM} & \mytextsf{Multimodal} & \mytextsf{SAN}\\
        \hline
        89.2 & 80.0 & 6.9 & 22.4 & 22.3 & 10.0\\
      \end{tabular}\\
      (a)\\
      \begin{tabular}{@{}l|c|c|c|c@{}}
        Ours/SA & \mytextsf{Standard} & \mytextsf{NC} & \mytextsf{NWNav} & \mytextsf{NWNavRec}\\
        \hline
        \texttt{nav\_obj} & 94.6/91.6 & 94.6/91.8 & 94.6/91.7 & 94.2/83.4\\
        \texttt{nav\_col\_obj} & 94.3/88.8 & 93.2/89.6 & 94.1/89.6 & 94.0/83.4\\
        \texttt{nav\_nr\_obj} & 93.1/74.5 & 92.7/74.2 & 93.0/77.2 & 93.0/70.8\\
        \texttt{nav\_bw\_obj} & 73.9/69.3 & 75.3/70.1 & 74.5/70.8 & 73.7/69.3\\
        \hline
        *\texttt{nav\_obj} & N/A & 93.8/90.6 & 94.5/88.3 & 94.7/4.1\\
        *\texttt{nav\_col\_obj} & N/A & 93.1/87.3 & 93.8/81.8 & 92.8/7.0\\
        *\texttt{nav\_nr\_obj} & N/A & 92.6/71.6 & 92.7/56.6 & 87.1/25.7\\
        *\texttt{nav\_bw\_obj} & N/A & 76.3/70.2 & 74.5/66.1 & 70.4/59.7\\
      \end{tabular}\\
      (b)\\
    \end{tabular}
  }
  \vspace{-1ex}
  \makeatletter
  \def\@captype{table}
  \makeatother
  \caption{Success rates (\%).
    (a) Overall rates of all the methods
    in a \mytextsf{Standard} training command setting.
    (b) Breakdown rates of our framework
    and \mytextsf{SimpleAttention} (SA) on the four subtasks under
    different training command conditions (columns).
    The last four rows show the rates of the test sessions that
    contain commands not seen in training.
  }
  \label{tab:rates}
  \vspace{-2ex}
\end{wrapfigure}

We use \mytextsf{SimpleAttention} for comparison as the other
baselines have far worse performance than it in the previous
experiment.
Both \mytextsf{SimpleAttention} and our framework are trained under
each condition without changing the their hyperparameters.
For testing, we put the held-out combinations/words back to the
commands (\ie\ \mytextsf{Standard} condition) and again test 10k
sessions.
Table~\ref{tab:rates}b contains the success rates.
Our framework gets almost the same success rates for all the
conditions, and obtains high zero-shot success rates.
The results of \mytextsf{NWNavRec} show that although some new object
concepts are learned from a completely different problem, they can be
tested on navigation without any model retraining or finetuning.
In contrast, \mytextsf{SimpleAttention} almost fails on zero-shot
commands (especially on \mytextsf{NWNavRec}).
Interestingly, in \texttt{nav\_bw\_obj} it has a much higher zero-shot
success rate than in the other subtasks.
Our analysis reveals that with the $3\times 3$ filter, it always
highlights the location between two objects without detecting their
classes, because usually there is only one qualified pair of objects
in the environment for \texttt{nav\_bw\_obj}.
As a result, it does not really generalize to unseen commands.

\begin{figure}[t!]
  \begin{center}
    \resizebox{\textwidth}{!}{
      \begin{tabular}{@{}llll@{}}
        \textit{Watermelon is the} &
        \textit{Reach the grid between} &
        \textit{Red ladybug is the} &
        \textit{Can you go to the}\\
        \textit{destination.} &
        \textit{grape and deer.} &
        \textit{target.} &
        \textit{south of the elephant?}\\\\
        \includegraphics[width=0.25\textwidth]{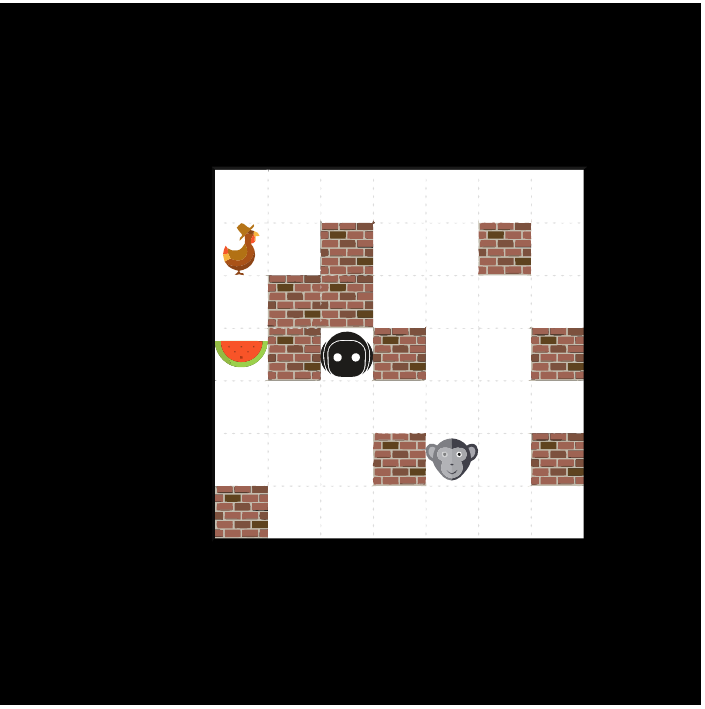} &
        \includegraphics[width=0.25\textwidth]{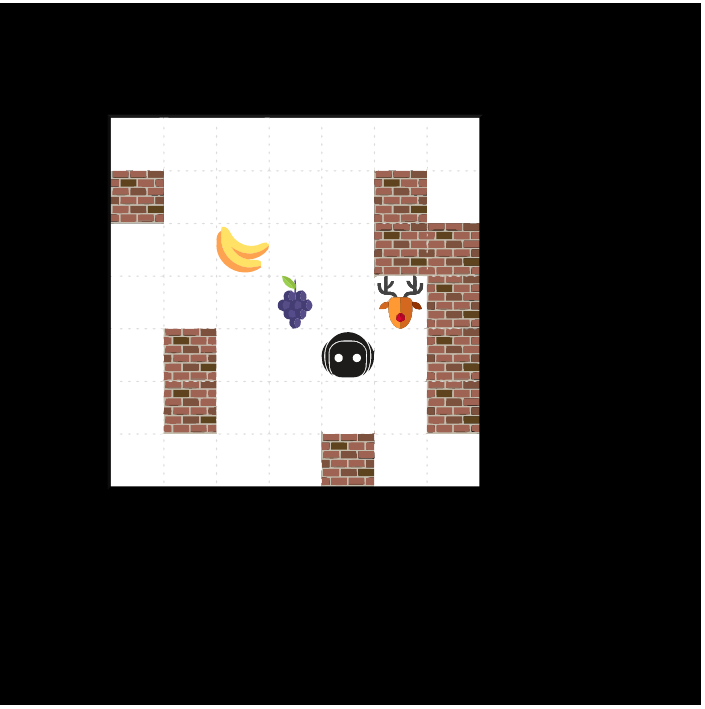} &
        \includegraphics[width=0.25\textwidth]{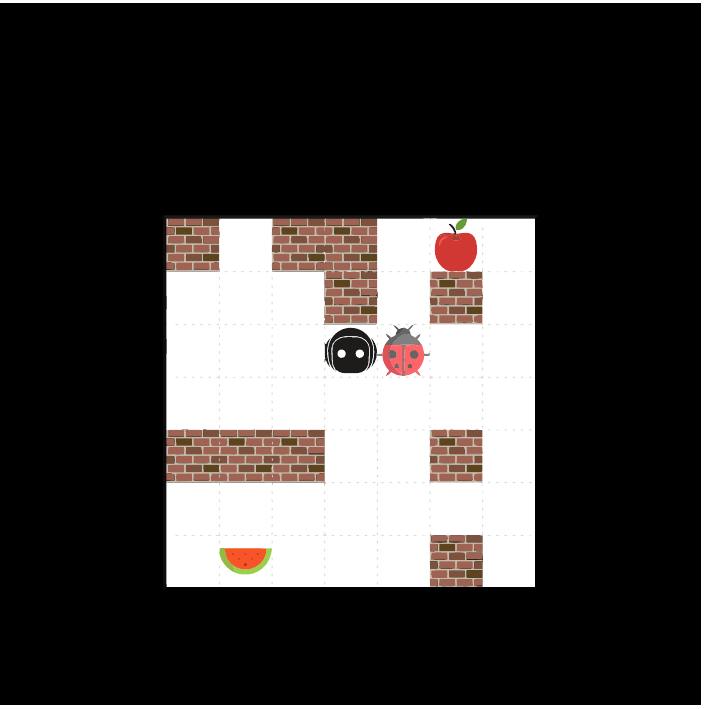} &
        \includegraphics[width=0.25\textwidth]{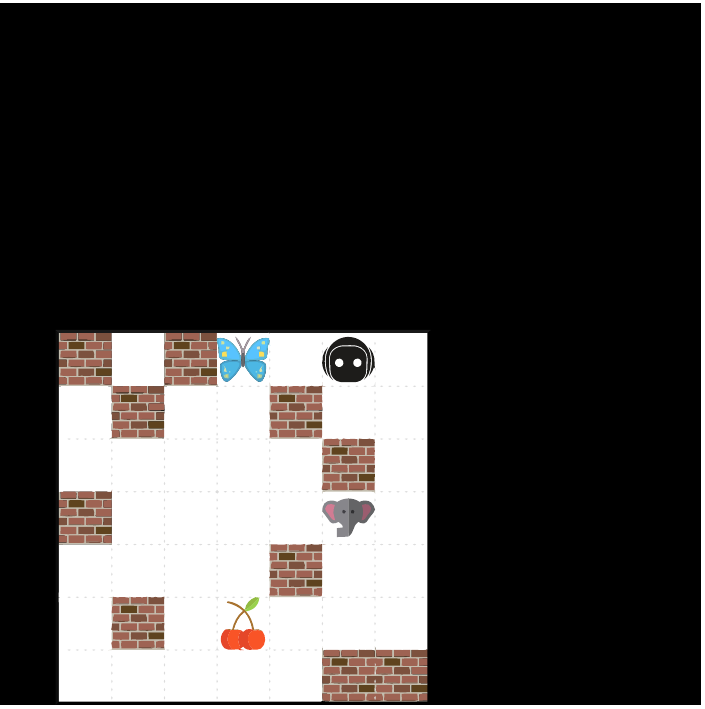}\\
        \includegraphics[width=0.25\textwidth]{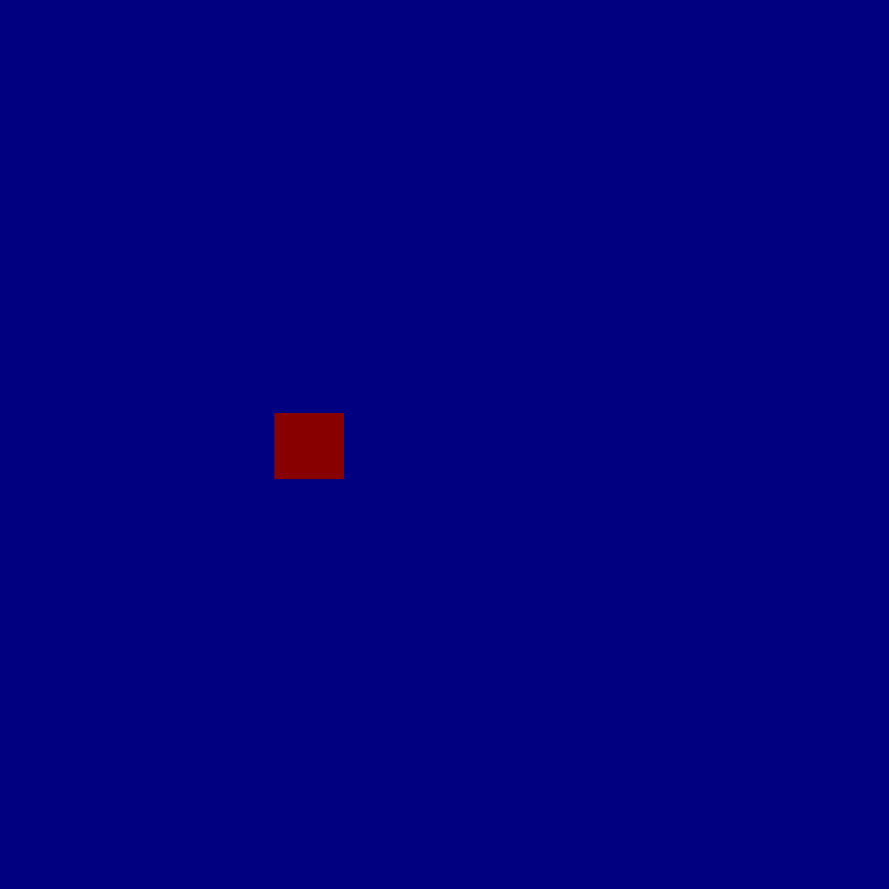} &
        \includegraphics[width=0.25\textwidth]{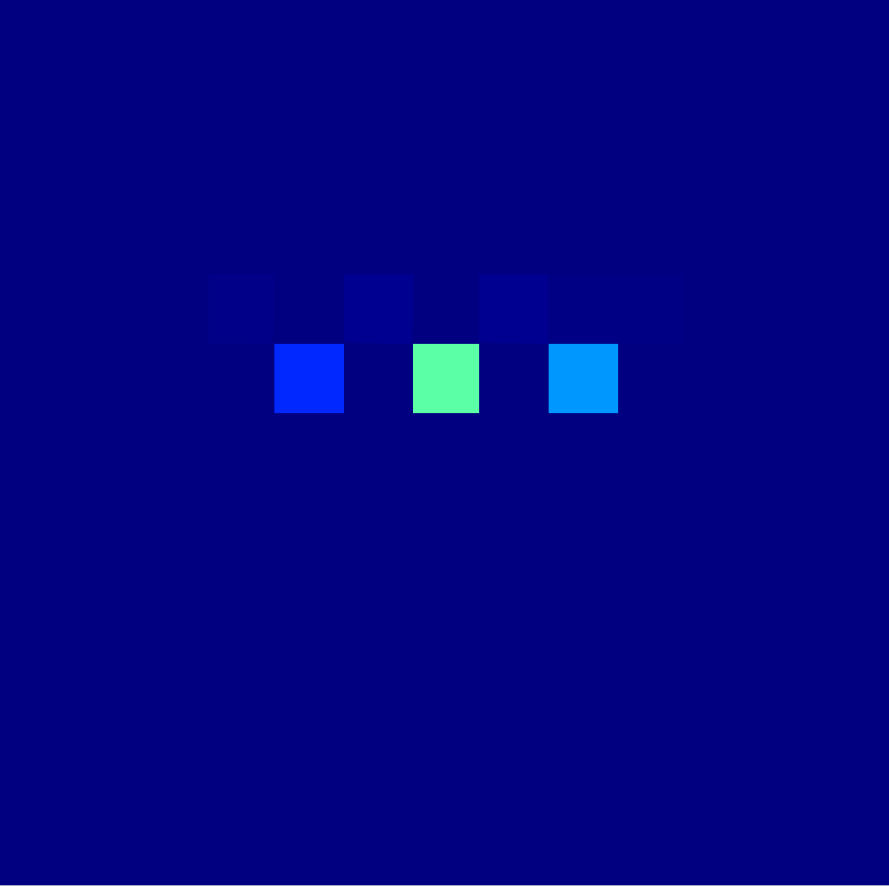} &
        \includegraphics[width=0.25\textwidth]{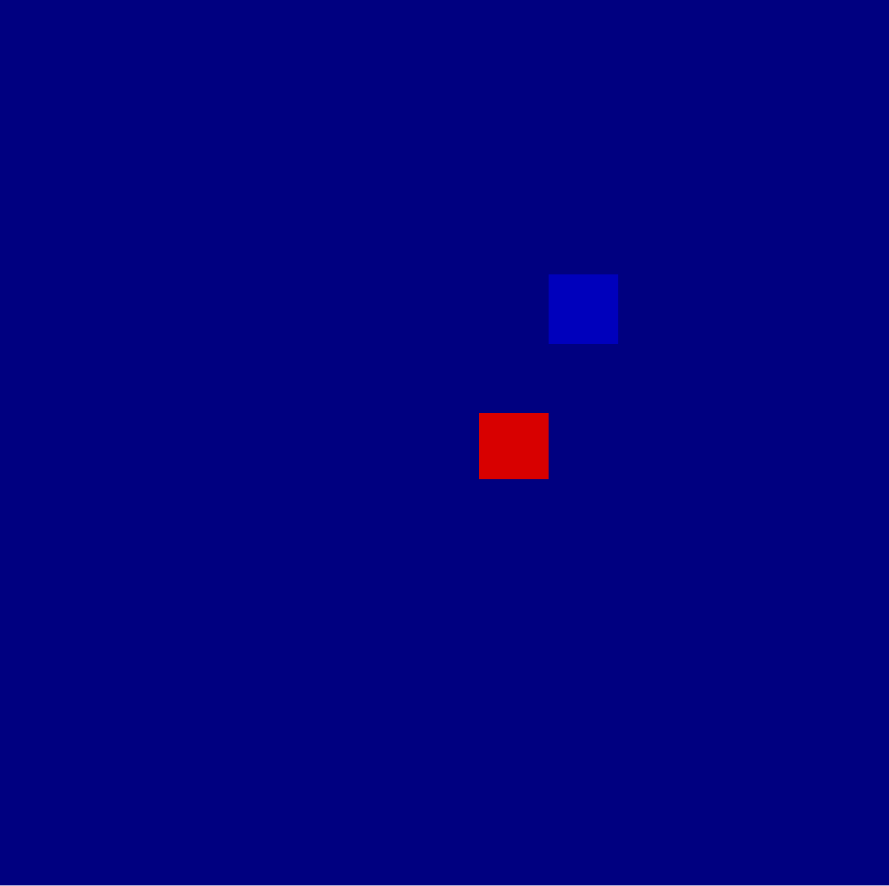} &
        \includegraphics[width=0.25\textwidth]{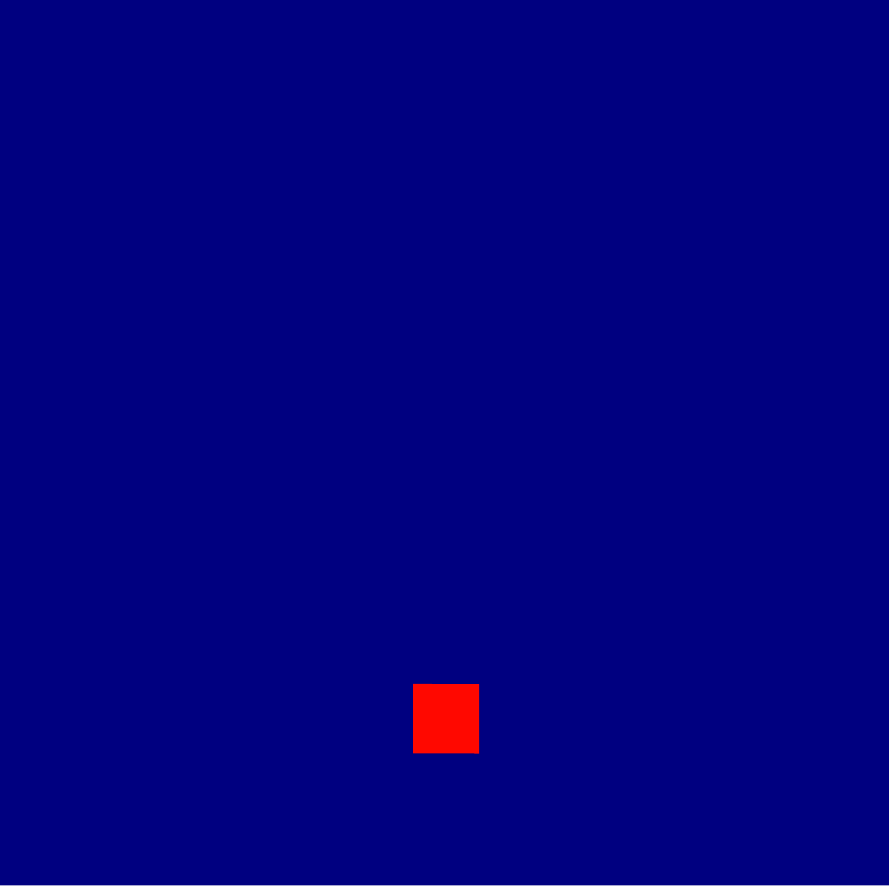}\\
      \end{tabular}
    }
    \caption{Examples of attention maps in different sessions.
      Top: the navigation commands.
      Middle: the current environment images.
      Bottom: the corresponding attention maps output by the language
      module.
      Note that attention maps are all egocentric: the map center is
      the agent's location.
    }
    \label{fig:example-att}
  \end{center}
\end{figure}

\textbf{Visualization and Analysis} Our framework produces
intermediate results that can be easily visualized and analyzed.
One example has been shown in Figure~\ref{fig:overview}, where the
environment map is produced by a trained perception module.
It detects exactly all the obstacles and goals.
The map constitutes the learned prior knowledge for navigation: all
walls and goals should be avoided by default because they incur
negative rewards.
This prior, combined with the attention map produced for a command,
contains all information the agent needs to navigate.
The attention maps are usually very precise
(Figure~\ref{fig:example-att} Appendix~\ref{app:figures}), with some
rare cases in which there are flaws, \eg\ when the agent needs to
navigate between two objects.
This is due to our simplified assumption on 2D geometric
translation: the attention map of a location word is treated as a
filter and the translation is modeled as convolution.
This results in attention diffusion in the above case.
To address this issue, more complicated transformation can be used
(\eg\ FC layers).

\begin{figure}[t!]
  \begin{center}
    \includegraphics[width=\textwidth]{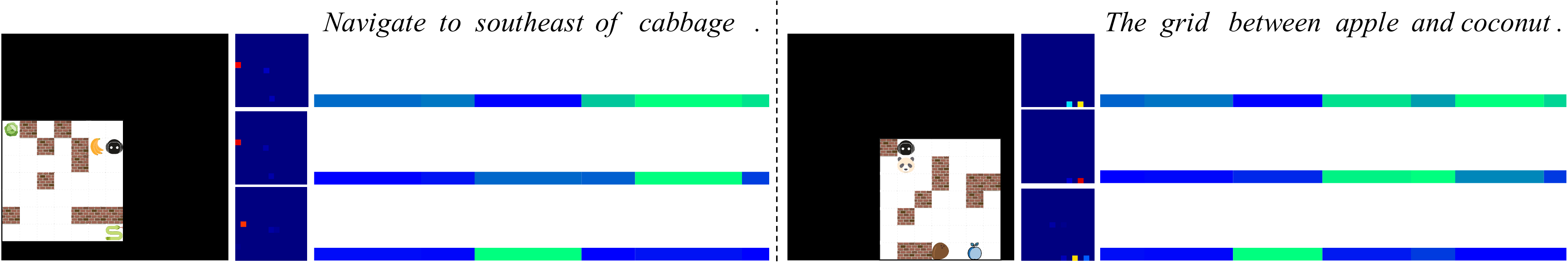}
    \caption{Illustration of the word attention on two examples.
      Given the current environment image and a navigation command,
      the programmer generates an attention map in three steps.
      At each step the programmer RNN focuses on a different portion
      of the sentence.
      The word attention is visualized by a color strip where
      brighter portion means more attention.
      On the left of each color strip is the corresponding attention
      map combining the current attention and the previously cached
      one (Figure~\ref{fig:overview} Right).
      The last attention map is used as the output of the programmer.
    }
    \label{fig:example-prog}
  \end{center}
\end{figure}

We also visualize the programming process.
We observe that the programmer RNN is able to shift its focus across
steps (Figure~\ref{fig:example-prog} Appendix~\ref{app:figures}).
With additional analysis of the grounding and map operations, we find
that in the first example, the programmer essentially does
\texttt{Transform}[\textit{southeast}](\texttt{Find}[\textit{cabbage}]($\mathbf{F}$)),
and in the second example, it essentially performs
\texttt{Transform}[\textit{between}](\texttt{CombineOr}(\texttt{Find}[\textit{apple}]($\mathbf{F}$),
\texttt{Find}[\textit{coconut}]($\mathbf{F}$))).

\begin{figure}[t!]
  \setlength{\tabcolsep}{1pt}
  \begin{center}
    \resizebox{\textwidth}{!}{
      \begin{tabular}{@{}ccc|ccc@{}}
        \multicolumn{3}{c|}{\textit{Could you please go to the dog?}} &
        \multicolumn{3}{c}{\textit{Please reach south of the cucumber.}}\\
        \includegraphics[width=0.16\textwidth]{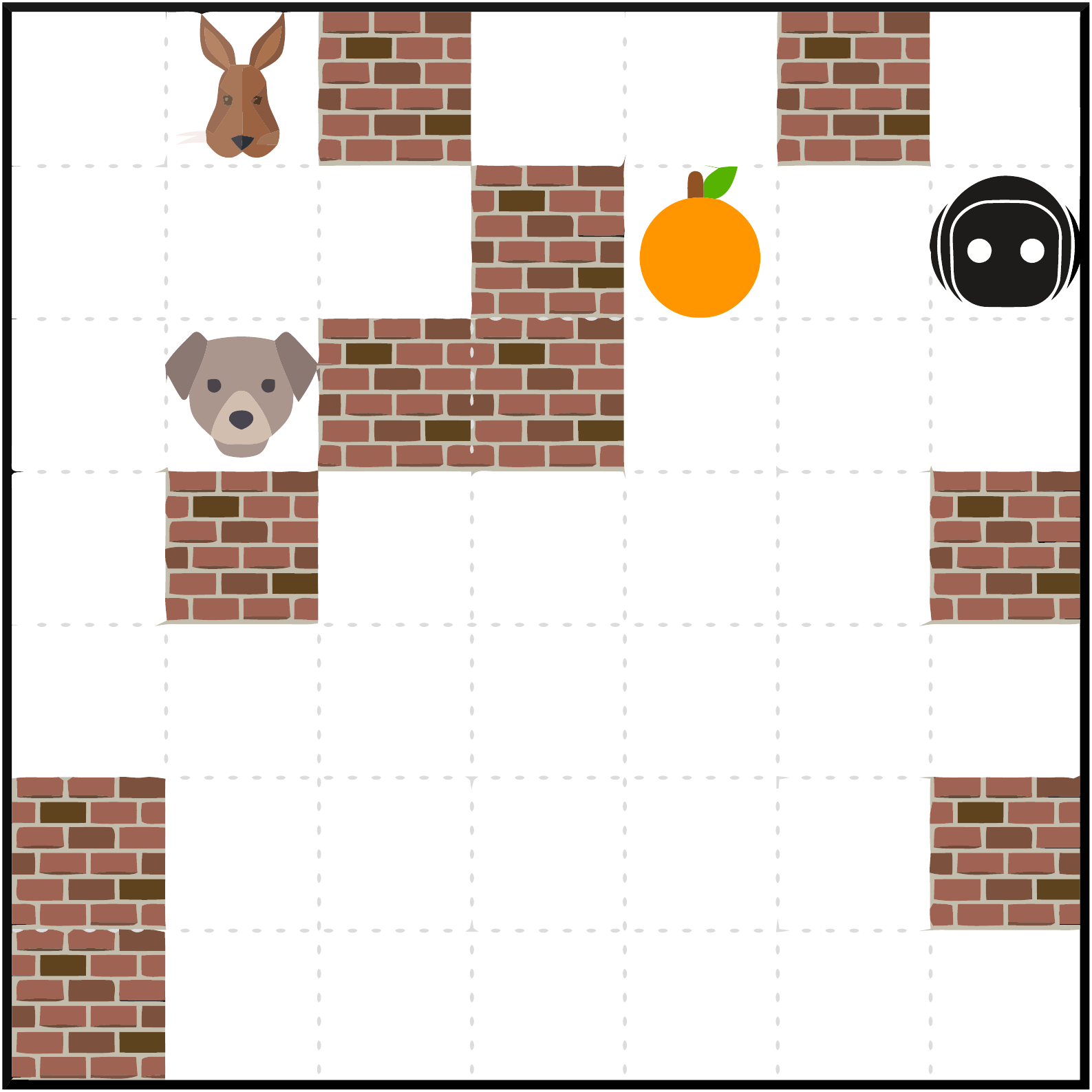} &
        \includegraphics[width=0.16\textwidth]{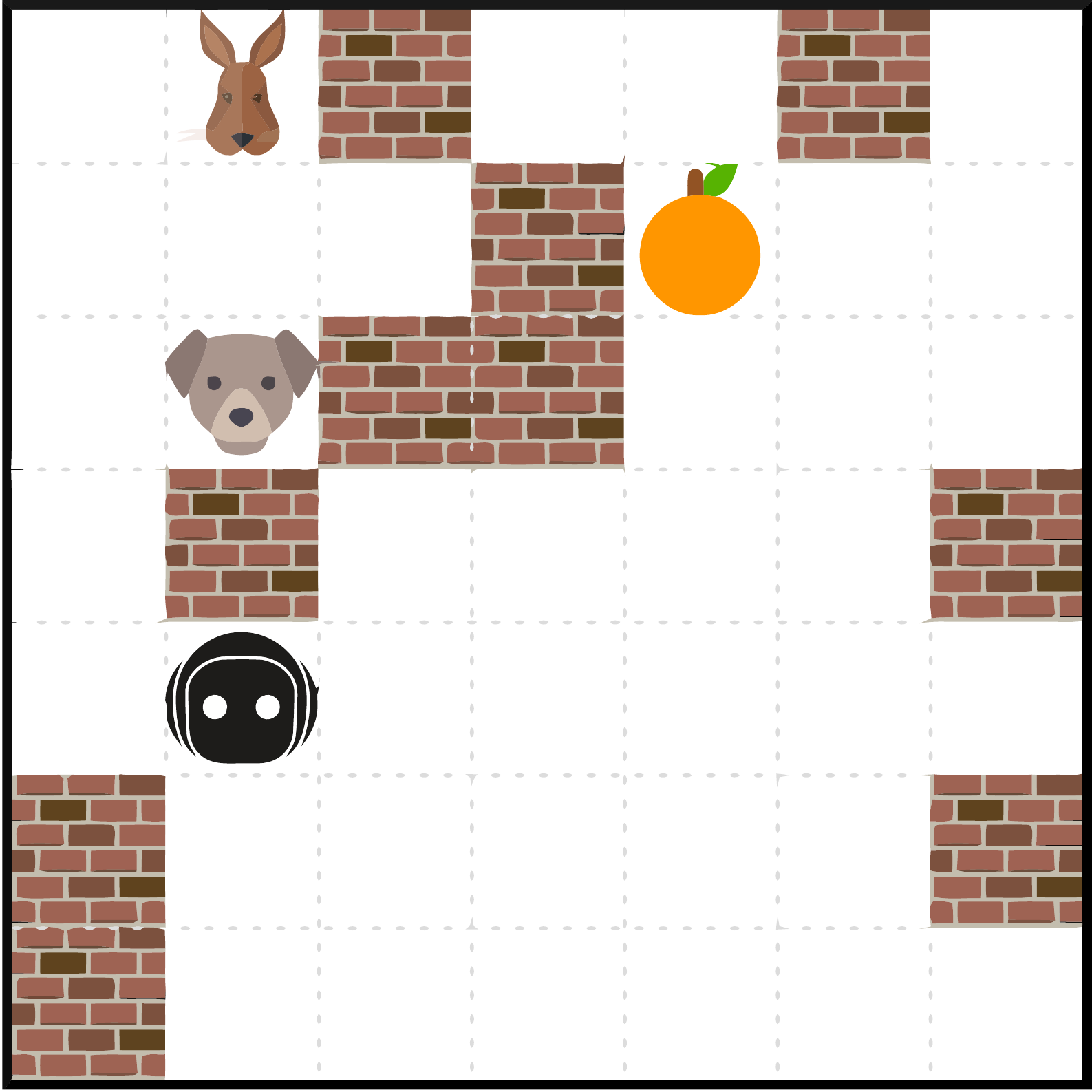} &
        \includegraphics[width=0.16\textwidth]{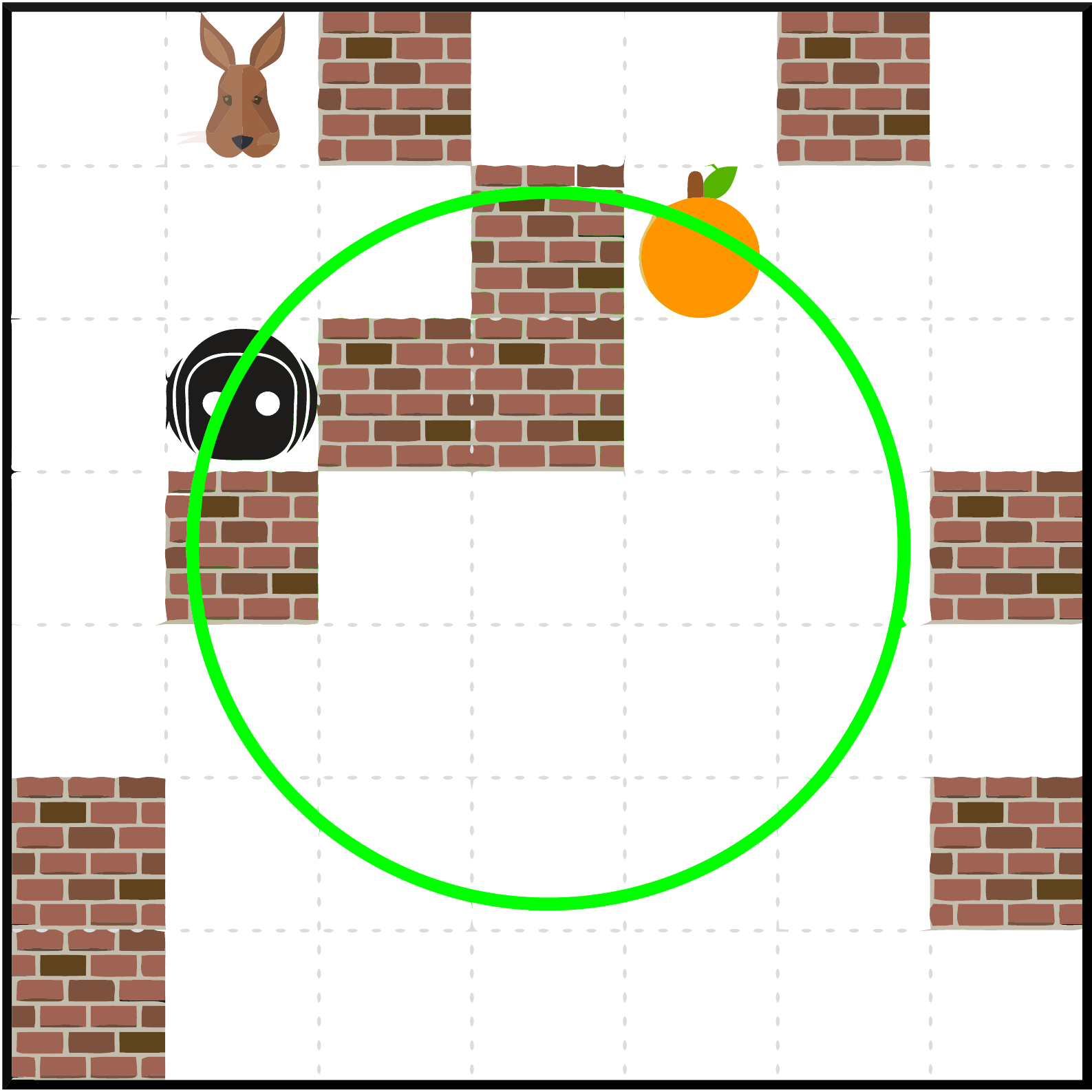} &
        \includegraphics[width=0.16\textwidth]{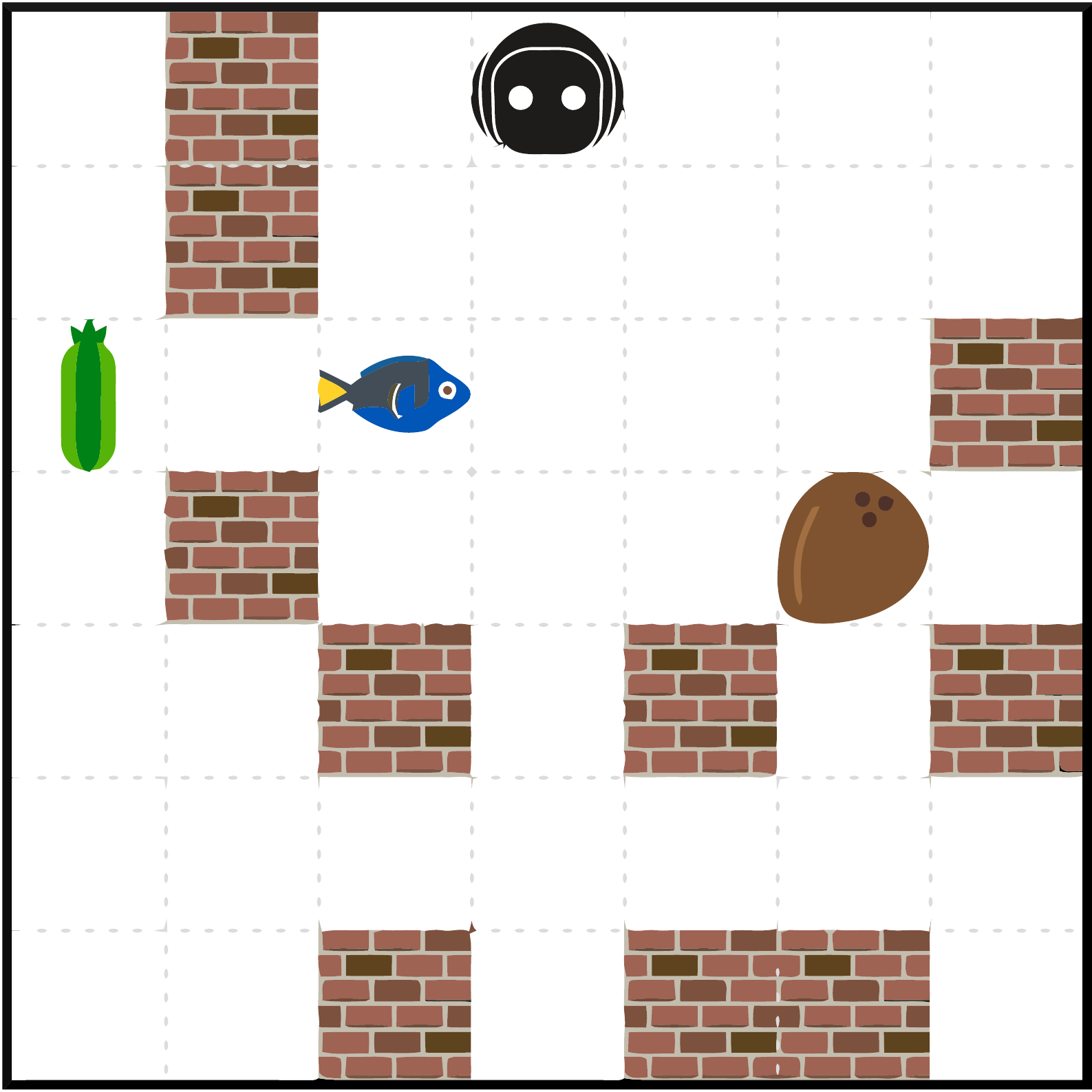} &
        \includegraphics[width=0.16\textwidth]{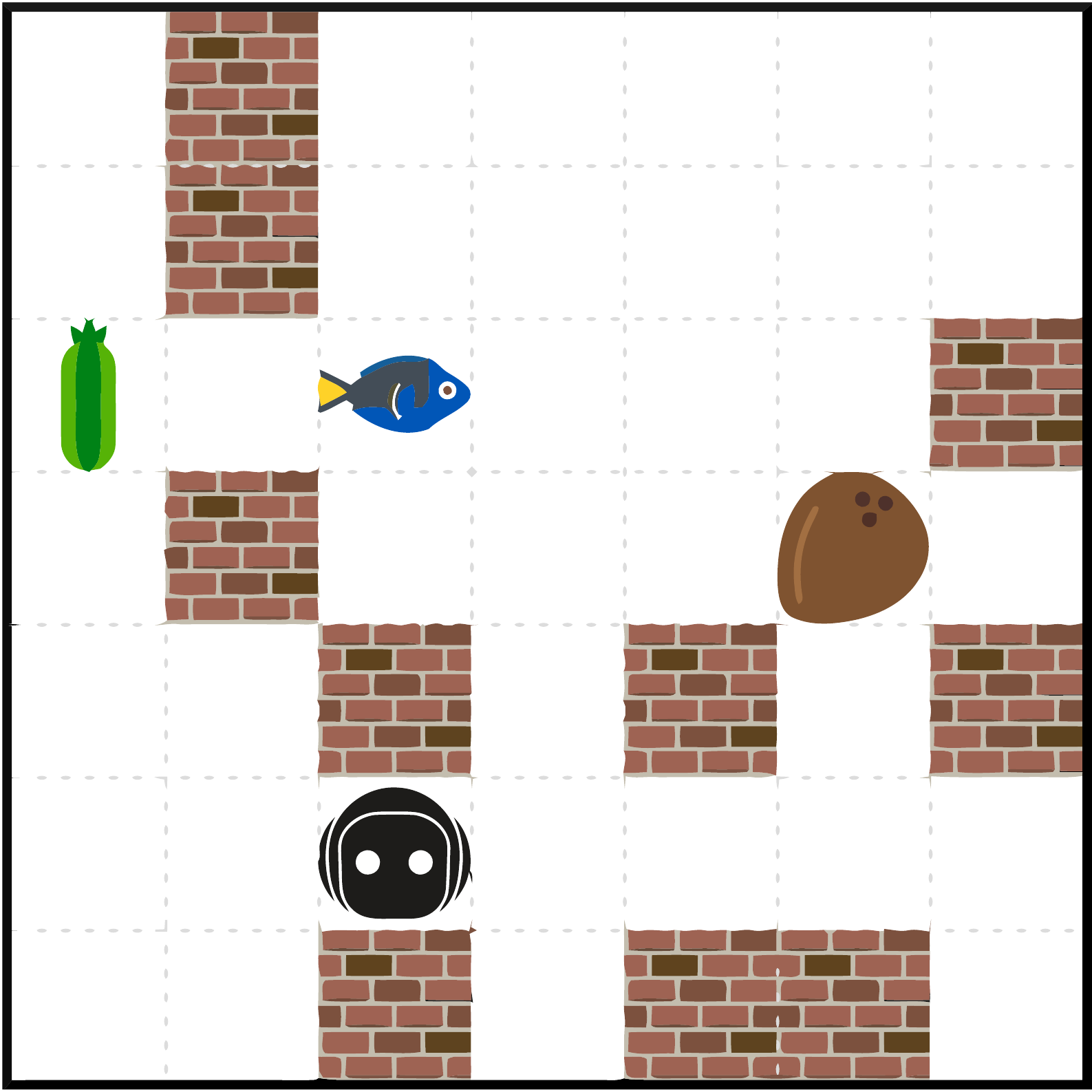} &
        \includegraphics[width=0.16\textwidth]{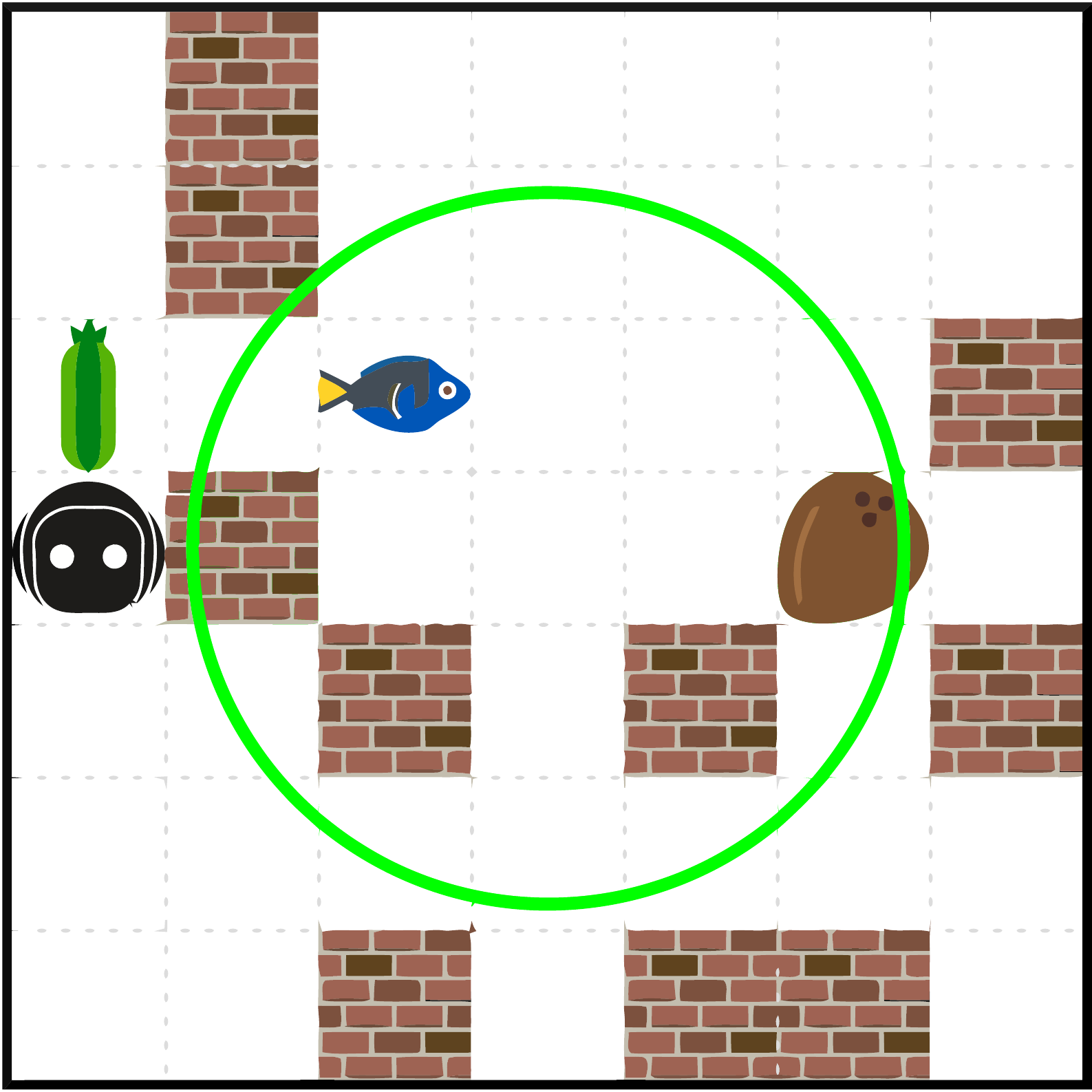}\\
      \end{tabular}
    }
    \caption{Examples of bypassing long walls.
      For each path, only three key steps are shown.
      (Green circles indicate successes.)
    }
    \label{fig:nav-examples}
  \end{center}
\end{figure}

We find the attention and environment maps very reliable by
visualization.
This is verified by the $\sim$100\% QA accuracy in recognition.
However, in Table~\ref{tab:rates} the best success rate is still
$\sim5\%$ away from the perfect.
Further analysis reveals that the agent tends to stuck in loop if the
target location is behind a long wall, although it has a certain
chance to bypass it (Figure~\ref{fig:nav-examples}
Appendix~\ref{app:figures}).
Thus we believe that the discrepancy between the good map quality and
the imperfect performance results from our action module.
Currently the action module learns a direct mapping from an
environment state to an action.
There is no support for either history remembering or route planning
\citep{Tamar2016}.
Since our focus here is zero-shot navigation, we leave such
improvements to future work.

\vspace{-3ex}
\section{Conclusion}
\vspace{-3ex}
We have demonstrated an end-to-end compositional framework for a
virtual agent to generalize an existing skill to new concepts without
being retrained.
Such generalization is made possible by reusing knowledge, encoded by
language and learned in other tasks.
By assembling words in different ways, the agent is able to tackle new
tasks while exploiting existing knowledge.
We reflect these ideas in the design of our framework and apply it to
a concrete example: zero-shot navigation in \textsc{xworld}.
Our framework is just one possible implementation.
Our claim is not that an intelligent agent must have a mental model as
the presented one, but it has to possess several crucial properties
discussed in Section~\ref{sec:framework}.
Currently the agent explores in a 2D environment.
In the future, we plan to migrate the agent to a 3D world like
Malmo \citep{Johnson2016}.
There will be several new challenges, \eg\ perception and geometric
transformation will be more difficult to model.
We hope that the current framework provides some preliminary insights
on training a similar agent in a 3D environment.

\section*{Acknowledgments}
%
%We thank professor Dhruv Batra for his valuable feedback on the manuscript.
%
We thank Yuanpeng Li, Liang Zhao, Yi Yang, Zihang Dai, Qing Sun,
Jianyu Wang, and Xiaochen Lian for their helpful suggestions on
writing.

{\small
\setlength{\bibsep}{0.1pt plus 0.3ex}
\bibliographystyle{abbrvnat}
\bibliography{robot_learning}
}

\clearpage
\section{Appendix}
\subsection{Word Attention}
\label{app:word-attention}
\begin{figure}[h!]
  \begin{center}
    \includegraphics[angle=90,width=0.95\textwidth]{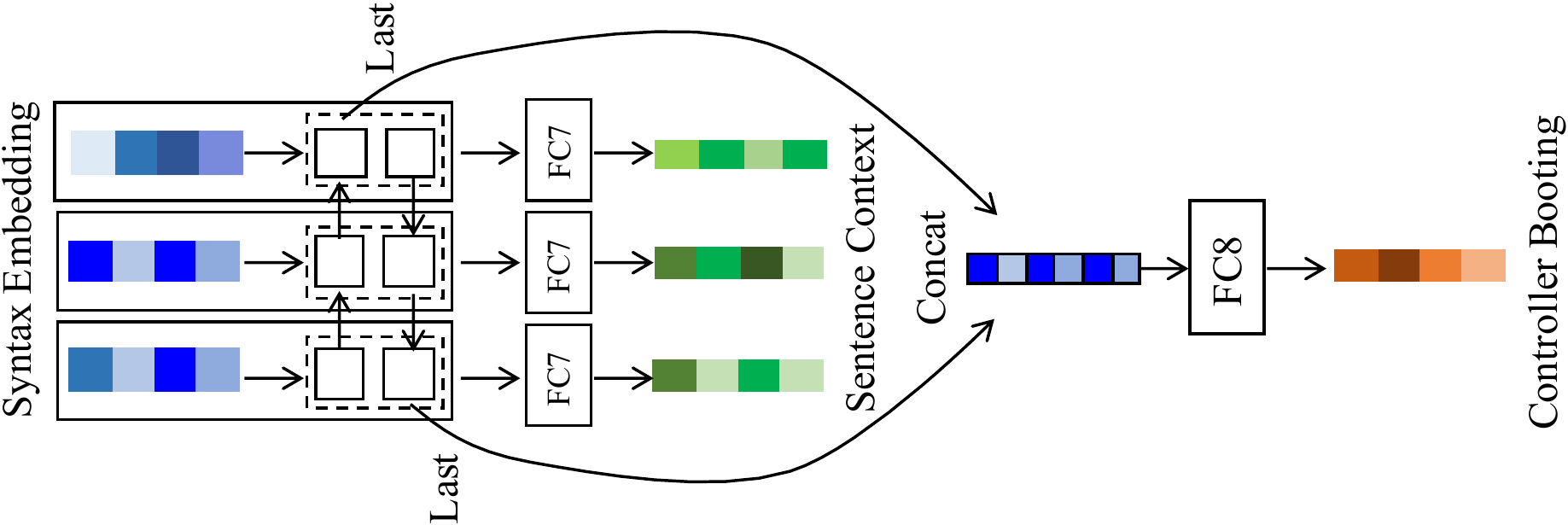}
    \caption{A Bidirectional RNN that receives a sequence of
      {\color{blue}syntax embeddings}, and outputs a sequence of
      {\color{ForestGreen}word contexts} and an
      {\color{orange}RNN booting} vector.
    }
    \label{fig:brnn}
  \end{center}
\end{figure}

A sentence of length $L$ is first converted to a sequence of word
embeddings $\mathbf{e}_l$ by looking up the embedding table
$\mathbf{E}$.
Then the embeddings are projected to syntax embeddings
(Figure~\ref{fig:projections}).
The syntax embeddings are fed into a Bidirectional RNN
\citep{Schuster1997} to obtain word context vectors $\mathbf{e}_l^c$.
The last forward state and the first backward state are concatenated
and projected to a booting vector (Figure~\ref{fig:brnn}), which is
used to initialize the programmer RNN.
Given a fixed number of programming steps, at each step $s$ the
programmer RNN computes the attention for each word $l$ from its
context vector $\mathbf{e}_l^c$:
\begin{equation*}
  \begin{array}{l}
    c_{s,l}=Softmax_L\left(CosSim
    \left(\mathbf{h}_s,g(\mathbf{W}\mathbf{e}_l^c+\mathbf{b})\right)\right),\\
    CosSim(\mathbf{z},\mathbf{z}')=\frac{\mathbf{z}^{\intercal}\mathbf{z}'}
         {\norm{\mathbf{z}}\norm{\mathbf{z}'}},
  \end{array}
\end{equation*}
where $\mathbf{W}$ and $\mathbf{b}$ are projection parameters,
$\mathbf{h}_s$ is the RNN state, $g$ is the activation function, and
$Softmax_L$ is the Softmax function over $L$ words.
Then the context vectors $\mathbf{e}_l^c$ are weighted averaged by the
attention and fed back to the RNN to tell it how to update its hidden
state.

\subsection{Programmer Operations}
\label{app:operations}
The embedding masks produced by $\Phi$ and $\Psi$ support switching
among the \texttt{DescribeColor}, \texttt{DescribeLocation}, and
\texttt{DescribeName} operations by masking word embeddings according
to given language tokens (Section~\ref{sec:recognition}).
The selective map combination in Figure~\ref{fig:overview} Right
supports nine different \texttt{TranslateAttention} operations by
convolution.
Our soft word attention supports the \texttt{CombineAnd} and
\texttt{CombineOr} operations by attending to multiple words
simultaneously.
For example, when multiple objects are attended, the resulting
attention map would be a union of the maps of the individual objects.
When an object and the color modifying that object are both attended,
the resulting attention map would be an intersection of the two maps,
thus selecting the object with the correct color from a set of objects
with the same name.
Thus compared to NMNs \citep{Andreas2016a}, we end up with an
implementation of an implicit and differentiable modular network.

\subsection{Implementation Details}
\label{app:implementation}
The agent at each time step receives a $156\times 156$ RGB image.
This image is egocentric and includes both the environment and the
black padding region.
The agent processes the input image with a CNN that has four
convolutional layers: $(3,3,64), (2,2,64), (2,2,512), (1,1,512)$,
where $(x,y,z)$ represents $z$ $x\times x$ filters with stride $y$.
All the four layers have the ReLU activation function.
The output is the visual feature map with $512$ channels.
We stack it along the channel dimension with another parametric
spatial feature map of the same sizes.
This spatial feature map is initialized with zero mean and standard
deviation (Figure~\ref{fig:overview}c).

\begin{figure}[h!]
  \begin{center}
    \includegraphics[width=0.4\textwidth]{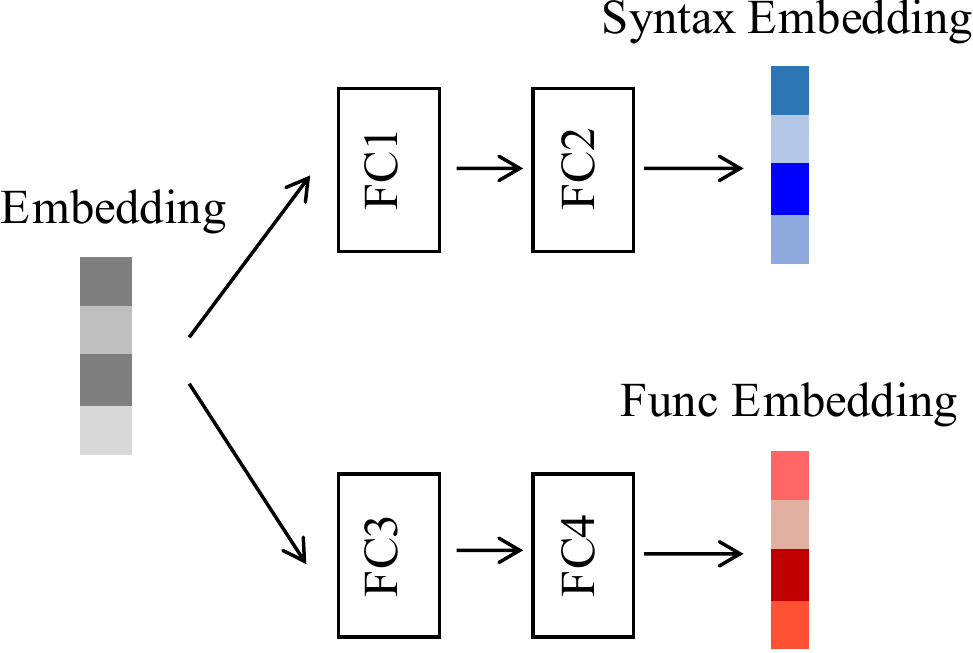}
    \caption{The projections from word embedding to
      {\color{blue}syntax embedding} and {\color{red}functionality
        embedding}.
    }
    \label{fig:projections}
  \end{center}
\end{figure}

\begin{figure}[h!]
  \begin{center}
    \begin{tabular}{@{}c|c@{}}
      \includegraphics[width=0.74\textwidth]{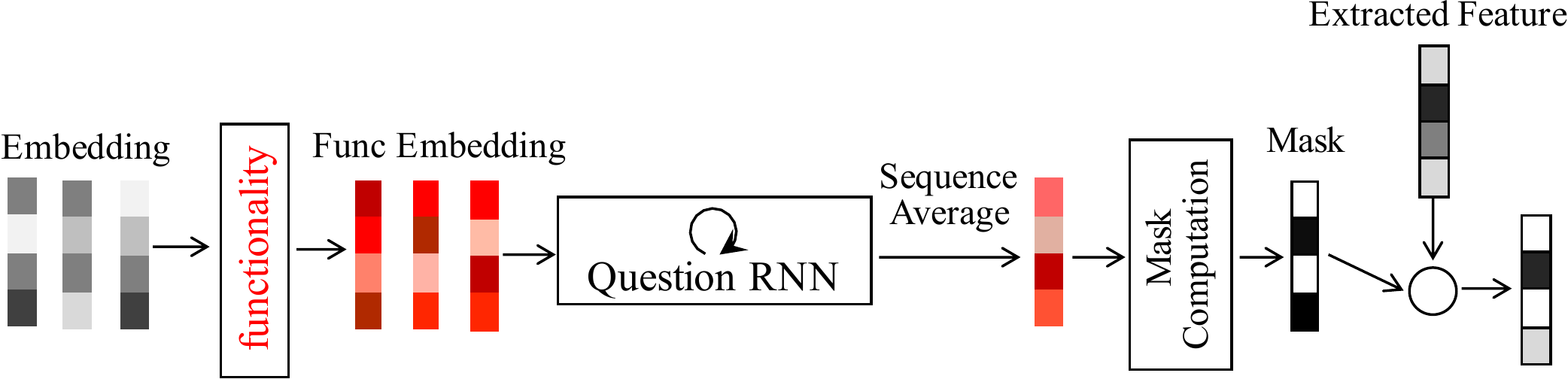} &
      \includegraphics[width=0.26\textwidth]{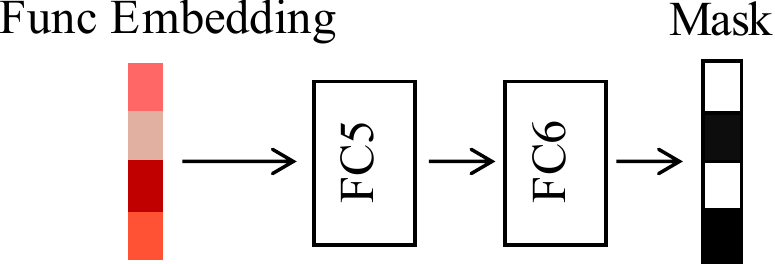}\\
      (a)&(b)\\
    \end{tabular}
    \caption{Details of computing the embedding masks.
      (a) The pipeline of Question Intention in
      Figure~\ref{fig:overview}b, which computes an embedding mask
      according to a question.
      (b) The two FC layers used by Mask Computation in both (a) and
      Figure~\ref{fig:overview} Right.
      They project a {\color{red}functionality embedding} to a mask.
    }
    \label{fig:masks}
  \end{center}
\end{figure}

The agent also receives a navigation command at the beginning of a
session.
The same command is repeated until the end of the session.
The agent may or may not receive a question at every time step.
The dimensions of the word embedding, syntax embedding, and
functionality embedding are $1024$, $128$, and $128$, respectively.
The word embedding table is initialized with zero mean and a standard
deviation of 1.
The hidden FC layers for computing the syntax and functionality
embeddings have $512$ units (Figure~\ref{fig:projections}).
The bidirectional RNN for computing word contexts has a state size
of $128$ in both directions.
The output RNN booting vector and word context also have a length
of $128$.
The state size of the programmer RNN is equal to the length of the
booting vector (Figure~\ref{fig:overview} Right).
The hidden FC layer for converting a functionality embedding to an
embedding mask has a size of $128$.
The RNN used for summarizing the question intention has $128$ states
(Figure~\ref{fig:masks}).
All FC layers and RNN states in the language and recognition module
use Tanh as the activation function.
The only exception is the FC layer that outputs the embedding mask
(Sigmoid).

In the action module, the action CNN for processing the attention map
and the environment map has two convolutional layers $(3,1,64)$ and
$(3,1,4)$, both with paddings of 1.
They are followed by three FC layers that all have $512$ units.
All five layers use the ReLU activation function.

\subsection{Baseline Models}
The language module of \mytextsf{SimpleAttention} has the word
embedding size of $1024$.
The RNN has the same size with the word embedding.
The FC layer that produces the $3\times 3$ filter has an output size
of $4608$ which is 9 times the channel number of the visual feature map.
The rest of the layer configuration is the same with our framework.
\mytextsf{VIS-LSTM} has a CNN with four convolutional layers $(3,2,64)$,
$(3,2,64)$, $(3,2,128)$, and $(3,1,128)$.
This is followed by three FC layers with size $1024$.
The word embedding and the RNN both have sizes of $1024$.
The RNN output goes through three FC hidden layers of size $512$
either for recognition or navigation.
\mytextsf{Multimodal} has the same layer size configuration with that
of \mytextsf{VIS-LSTM}.
\mytextsf{SAN} has a CNN with four convolutional layers $(3,3,64)$,
$(2,2,64)$, $(2,2,512)$, and $(3,1,1024)$, where the last layer makes
each grid have a $3\times3$ receptive field.
Its word embedding size and RNN state size are both $512$.
Following \citet{Yang2016}, we use two attention layers.

The outputs of all layers of the above baselines are ReLU activated
except for those that are designed as linearly activated.

\subsection{\textsc{xworld} Setup}
\label{app:xworld}
We configure square environments with sizes ranging from 3 to 7.
We fix the size of the environment image by padding walls for smaller
environments.
Different sessions may have different map sizes.
In each session,
\begin{compactenum}[$\circ$]
  \item The number of time steps $T$ is four times the map size.
  \item The number of objects on the map is from 1 to 3.
  \item The number of wall grids on the map is from 0 to 10.
  \item The positive reward when the agent reaches the correct
    location is set to $1$.
    The negative rewards for hitting walls and for stepping
    on non-target objects are set to $-0.2$ and $-1$, respectively.
    The time step penalty is set to $-0.1$.
\end{compactenum}

The teacher has a vocabulary size of 104, including 2 punctuation
marks.
There are 9 locations, 4 colors, and 40 distinct object classes.
Each object class has $2.85$ object instances on average.
Every time the environment is reset, a number of object classes are
randomly sampled and an object instance is randomly sampled for each
class.
There are in total 16 types of sentences the teacher can speak,
including 4 types of navigation commands and 12 types of recognition
questions.
Each sentence type has multiple non-recursive natural-language
templates, and corresponds to a subtask the agent must learn to
perform.
In total there are 256,832 distinct sentences with 92,442 for the
navigation task and 164,390 for the recognition task.
The sentence length ranges from 2 to 12.

The object, location, and color words of the teacher's language are
listed below.
These are the content words with actual meanings that can be grounded
in the environment.
All the other words are treated as grammatical words whose embeddings
are only for interpreting sentence structures.
The differentiation between content and grammatical words is
automatically learned by the agent based on the teacher's language and
the environment.
All words have the same form of representation.

\begin{center}
  \resizebox{\textwidth}{!}{
    \begin{tabular}{llll}
      Object & Location & Color & Other\\
      \hline\\
      \textit{apple, avocado, banana, blueberry, butterfly,} &
      \textit{east, west,} & \textit{green,} & \textit{?, ., and,
        block, by, can, color, could,}\\
      \textit{cabbage, cat, cherry, circle, coconut,} & \textit{north,
        south,} & \textit{red,} & \textit{destination, direction,
        does, find, go,}\\
      \textit{cucumber, deer, dog, elephant, fig,} & \textit{northeast,
        northwest,} & \textit{blue,} & \textit{goal, grid, have,
        identify, in, is, locate,}\\
      \textit{fish, frog, grape, hedgehog, ladybug,} &
      \textit{southeast, southwest,} & \textit{yellow} &
      \textit{located, location, me, move, name,}\\
      \textit{lemon, lion, monkey, octopus, orange,} &
      \textit{between} & & \textit{navigate, near, nothing, object,
        of, on,}\\
      \textit{owl, panda, penguin, pineapple, pumpkin,} & & &
      \textit{one, OOV, please, property, reach, say,}\\
      \textit{rabbit, snake, square, squirrel, star,} & & &
      \textit{side, target, tell, the, thing, three, to,}\\
      \textit{strawberry, triangle, turkey, turtle, watermelon} & & &
      \textit{two, what, where, which, will, you, your}\\
    \end{tabular}
  }
\end{center}

The sentence types that the teacher can speak are listed below.
Each sentence type corresponds to a subtask.
The triggering condition describes when the teacher says that type of
sentences.
Besides the conditions shown, an extra condition for navigation
commands is that the target location must be reachable from the
current agent location.
An extra condition for color-related questions is that the object
color must be one of the four defined colors, and objects with other
colors will be ignored in these questions.
If at a time step there are multiple conditions triggered, we randomly
sample one sentence type for navigation and another for recognition.
After the sentence type is sampled, we generate the command or
question according to the corresponding sentence templates.

\vspace{-2ex}
\begin{center}
\resizebox{\textwidth}{!}{
\begin{tabular}{lll}
  Sentence Type & Example & Triggering Condition\\
  (Subtask) & &\\
  \hline\\
  \texttt{nav\_obj} & \textit{Please go to the apple.}
  & [C0] Beginning of a session. \&\\
  & & [C1] The referred object has a unique\\
  & & name.\\
  \texttt{nav\_col\_obj} & \textit{Could you please move to the
    red apple?} &
  [C0] \& [C2] There are multiple objects\\
  & & that either have the same name\\
  & & but different colors, or have different\\
  & & names but the same color.\\
  \texttt{nav\_nr\_obj} & \textit{The north of the apple is your
    destination.} & [C0] \& [C1]\\
  \texttt{nav\_bw\_obj} & \textit{Navigate to the grid
    between apple and} &
  [C0] \& [C3] Both referred objects have\\
  & \textit{banana please.} & unique names and are separated by\\
  & & one grid.\\
  & &\\
  \texttt{rec\_col2obj} & \textit{What is the red object?}
  & [C4] There is only one object that\\
  & & has the referred color.\\
  \texttt{rec\_obj2col} & \textit{What is the color of the
    apple?} & [C1]\\
  \texttt{rec\_loc2obj} & \textit{Please tell the name of the
    object in the south.} & [C5] The agent is near the referred\\
  & & object.\\
  \texttt{rec\_obj2loc} & \textit{What is the location of the
    apple?} & [C1] \& [C5]\\
  \texttt{rec\_loc2col} & \textit{What color does the object in
    the east have?} & [C5]\\
  \texttt{rec\_col2loc} & \textit{Where is the red object
    located?} & [C4] \& [C5]\\
  \texttt{rec\_loc\_obj2obj} & \textit{Identify the
    object which is in the east of the apple.}
  & [C1] \& [C6] The referred object is\\
  & & near another object\\
  \texttt{rec\_loc\_obj2col} & \textit{What is the
    color of the east to the apple?} & [C1] \& [C6]\\
  \texttt{rec\_col\_obj2loc} & \textit{Where is the
    red apple?} & [C2] \& [C5]\\
  \texttt{rec\_bw\_obj2obj} & \textit{What is the
    object between apple and banana?} &
  [C7] Both referred objects have\\
  & & unique names and are separated by\\
  & & one object.\\
  \texttt{rec\_bw\_obj2loc} & \textit{Where is the
    object between apple and banana?}
  & [C7] \& [C8] The agent is near the \\
  & & object in the middle.\\
  \texttt{rec\_bw\_obj2col} & \textit{What is the
    color of the object between apple} & [C7]\\
  & \textit{and banana?} &\\
\end{tabular}
}
\end{center}

\subsection{Experience Replay and Curriculum Learning}
\label{app:training}
We employ Experience Replay \citep{Mnih2015} for training both the
navigation and recognition tasks.
The environment inputs, rewards, and the actions taken by the agent at
the most recent 10k time steps are stored in a replay buffer.
During training, every time two minibatches of the same number of
experiences are sampled from the buffer, one for computing
$\nabla_{\theta}\mathcal{L}_{SL}(\theta)$ and the other for computing
$\nabla_{\theta}\mathcal{L}_{RL}(\theta)$.
For the former, only individual experiences are sampled.
We uniformly sample experiences from a subset of the buffer which
contains the teacher's questions.
For the latter, we need to sample transitions (\ie\ pairs of
experiences) so that TD error can be computed.
We sample from the entire buffer using the Rank-based Sampler
\citep{Schaul2015} which has proven to increase the learning
efficiency by prioritizing rare experiences in the buffer.

Because in the beginning the language is quite ambiguous, it is
difficult for the agent to start learning with a complex environment
setup.
Thus we exploit Curriculum Learning \citep{Bengio2009} to gradually
increase the environment complexity.
We gradually change the following things linearly in proportional to
$\min(1, G' \mathbin{/} G)$, where $G'$ is the number of sessions so
far and $G$ is the number of curriculum sessions:
\begin{compactenum}[$\circ$]
\item The number of grids of the environment.
\item The number of objects in the environment.
\item The number of wall grids.
\item The number of possible object classes that can appear in the environment.
\item The length of a navigation command or a recognition question.
\end{compactenum}
We find that this curriculum is crucial for efficient learning, because
in the early phase the agent is able to quickly master the meanings of
the location and color words given only small ambiguity.
After this, these words are used to guide the optimization when more
and more new sentence structures and objects are added.
In the experiments, we set $G=10\text{k}$ during training while do not
use any curriculum during test (maximal difficulty).

\clearpage
\subsection{Figures}
\label{app:figures}

\clearpage

\clearpage

\end{document}